\documentclass{article}
\usepackage{booktabs}
\usepackage{color}
\usepackage{caption}
\usepackage[preprint]{spconfa4}
\usepackage{amsmath,epsfig}
\usepackage{amssymb}
\usepackage{subcaption}
\usepackage{ulem}
\usepackage{graphicx}
\graphicspath{ {./figures/} }
\graphicspath{ {./_images/} }

\let\OLDthebibliography\thebibliography
\renewcommand\thebibliography[1]{
	\OLDthebibliography{#1}
	\setlength{\parskip}{0pt}
	\setlength{\itemsep}{0pt plus 0.3ex}
}

\makeatletter
\renewcommand\normalsize{%
	\@setfontsize\normalsize\@xpt\@xiipt
	\abovedisplayskip 4\p@ \@plus1\p@ \@minus5\p@
	\abovedisplayshortskip \z@ \@plus1\p@
	\belowdisplayshortskip 7\p@ \@plus1\p@ \@minus1\p@
	\belowdisplayskip \abovedisplayskip
	\let\@listi\@listI}

\begin{document}\sloppy
	
	% Example definitions.
	% --------------------
	\def\x{{\mathbf x}}
	\def\L{{\cal L}}

	% Title.
	% ------
	\title{Invertible Network for Unpaired Low-Light Image Enhancement}
	%
	% Single address.
	% ---------------
	\name{Jize Zhang, Haolin Wang, Xiaohe Wu and Wangmeng Zuo}
	\address{School of Computer Science and Technology, Harbin Institute of Technology, China \\ 
	 jize.zhang.cs@outlook.com,
	 Why\_cs@outlook.com,
	 xhwu.cpsl.hit@gmail.com,
	 cswmzuo@gmail.com}

	\maketitle

	\begin{abstract}
		Existing unpaired low-light image enhancement approaches prefer to employ the two-way GAN framework, in which two CNN generators are deployed for enhancement and degradation separately. However, such data-driven models ignore the inherent characteristics of transformation between the low and normal light images, leading to unstable training and artifacts. Here, we propose to leverage the invertible network to enhance low-light image in forward process and degrade the normal-light one inversely with unpaired learning. The generated and real images are then fed into discriminators for adversarial learning. In addition to the adversarial loss, we design various loss functions to ensure the stability of training and preserve more image details. Particularly, a reversibility loss is introduced to alleviate the over-exposure problem. Moreover, we present a progressive self-guided enhancement process for low-light images and achieve favorable performance against the SOTAs.
	\end{abstract}

	\begin{figure*}[t]
		\centering
		\includegraphics[width=0.95\textwidth]{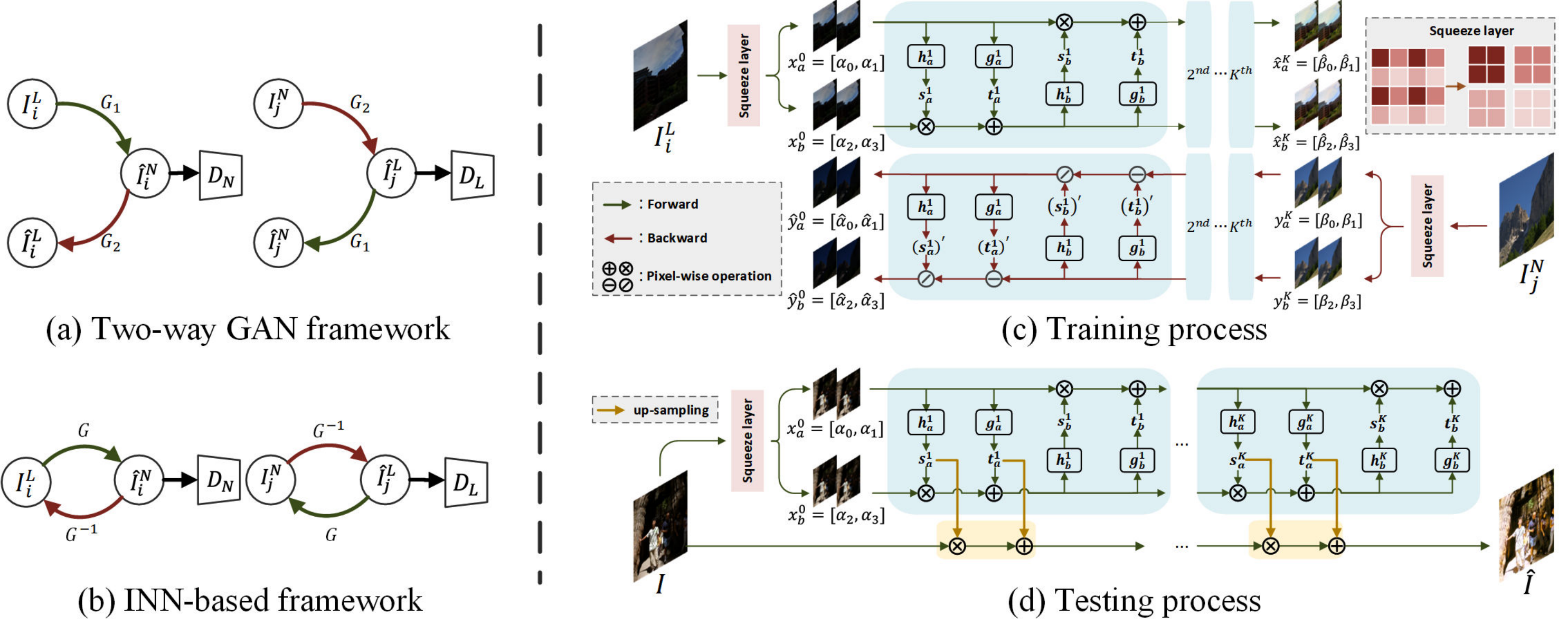}
		\vspace{-2mm}
		\caption{Overview of our Inv-EnNet. (a) and (b) illustrate the difference with the traditional two-way GAN framework. (c) shows the training process and (d) depicts the proposed progressive self-guided enhancement for testing.}
		\label{fig:framework}
	\end{figure*}
	\section{Introduction}
	\label{sec:intro}
	High quality images are of great importance in various computer vision applications.
	Unfortunately, images captured in low-light environments usually suffer from unsatisfied visibility, such as low contrast.
	Therefore, low-light Image Enhancement (LIE) remains a challenge, especially due to the details in under-exposed regions are usually imperceptible.
	Traditional methods often enhance the low-light images by adjusting its contrast with global operator based on histogram equalization (HE)~\cite{he-10.1109/TCE.2007.4429280, he-article} or gamma correction~\cite{Huang2013EfficientCE}.
	Considering the insufficient illumination in low-light environments, Retinex methods~\cite{NPE, Fu_2016_CVPR, Guo2017LIMELI} propose to extract the illumination and reflection components. Various hand-crafted constraints and priors are designed in optimization models to recover or enhance the illumination of the image.
	However, these classical approaches are limited in model capacity and cannot describe the physical model of low-light image accurately, thus are difficult to deal with challenging inputs and cannot produce satisfactory results.
	In recent years, deep learning based methods have been explored extensively for low-light image enhancement. Prior works mainly rely on the paired data for supervised learning~\cite{Chen2018Retinex, Wang_2019_CVPR, kind++, liu2021ruas}.
	However, the simulated paired data cannot describe the characteristics of the real low-light images well and are prone to result in artifacts or color distortion.
	While the paired training set from real scenes is cost-expensive to acquire and may suffer from misalignment issue.
	Consequently, the generalization ability of the model with paired supervision inevitably degrades and makes it intractable to handle the variety and complexity of the input low-light images.

	As a remedy, numerous efforts have been made on unsupervised learning for low light enhancement.
	Based on unpaired training set, GAN-based methods provide an intuitive solution to avoid the dependence on paired data effectively and achieve promising results in the general image enhancement.
	For example, both DPE~\cite{Chen_2018_CVPR_DPE} and QAGAN~\cite{QAGAN} adopt the two-way GAN~\cite{CycleGAN2017} framework to implement the user-oriented image enhancement with unpaired training set.
	GEN\&LEN~\cite{GleNet} even proposes a two-stage training scheme for unpaired image enhancement with two-way GAN.
	However, above methods are not specifically designed for low-light images, thereby showing limited performance on extremely dark pictures.
	EnlightenGAN~\cite{jiang2021enlightengan} is the first to successfully apply the unpaired learning for low-light image enhancement. It employs an one way GAN structure to stabilize training and achieves considerable results.
	In the meanwhile, Guo et. al.~\cite{Zero-DCE} recently propose to train a lightweight CNN to estimate the image-specific and pixel-wise curves for dynamic range adjustment without the requirement of any paired or unpaired data.
	However, both of the models suffer from severe artifacts, while EnlightenGAN also produce over-exposed results.
	Two-way GAN based unpaired learning method are typical data-driven models relied on the elaborately designed discriminators, loss functions and the selection of unpaired data, ignoring the inherent characteristics of the transformation between the low and normal light images.
	{In practice, illumination variation can produce normal and low light images in a certain scene.}
	Therefore, it is more reasonable to learn a bijective function for modeling the special translation between the low and normal light images, which inspires us to adopt the invertible models.
	Using the unpaired training set, here, we propose an invertible network based framework for low-light image enhancement.
	In the forward process, the low-light image is enhanced to a normal-light photograph and the unpaired normal-light image is degraded to the corresponding low-light sample inversely.
	Then the generated and real images are leveraged to train different discriminators for adversarial learning.
	In the training phase, in addition to the adversarial loss, we design the transformation-consistent loss for stable training, the detail-preserving loss to maintain details of low light image, and the reversibility loss to alleviate the over-exposure problem.
	To prevent checkerboard artifacts caused by squeeze layers, we also present a progressive self-guided enhancement for inference and achieve the state-of-the-art performance.
	\section{Proposed method}
	\label{sec:meth}
	\subsection{Framework of Inv-EnNet}
	\label{sec:2.1}
	Under the unpaired learning setting, training set can be given by two independent sets of low-light images $\mathcal{X}^{L}$ and normal-light images $\mathcal{X}^{N}$. Here, $\mathbf{I}^{L}_{i}$ denotes a low-light image from $\mathcal{X}^{L}$ and $\mathbf{I}_{j}^{N}$ a normal-light image from $\mathcal{X}^{N}$.
	Different from the traditional two-way GAN framework in Figure~\ref{fig:framework}(a), we adopt the invertible network to model the reversibility translation between the low-light and normal-light images as shown in Figure ~\ref{fig:framework}(b).
	In particular, the detailed deployment of the invertible network architecture, namely Inv-EnNet, is illustrated in Figure~\ref{fig:framework}(c).
	In the forward process, taking the low-light image $\mathbf{I}^{L}_{i}$ as input, the goal is to enhance it and achieve the enhanced image. Conversely, the Inv-EnNet takes the unpaired normal-light image $\mathbf{I}^{N}_{j}$ as input and degrades it to generate a synthetic low-light image.
	Splitting the RGB channels into two parts directly would lead to poor information coupling due to the unique feature of each channel, and inevitably degrade the enhancement performance.
	In the meanwhile, the widely adopted wavelet down-sampling operation holds the same problem as it splits the image into low and high frequency information.
	Besides, it cannot be combined with our proposed enhancement method in Section~\ref{sec:2.2}.
	Inspired by RealNVP~\cite{realnvp}, we follow the spatial checkerboard pattern and adopt the squeeze layer to halve the resolution of images by reshaping the spatial $2\times2$ region into channel dimension.
	Typically, $\mathbf{I}^{L}_{i}$ is squeezed to four sub-images denoted as $\{\alpha_{i}\}_{i=0}^{3}$ and $\mathbf{I}^{N}_{j}$ is split into $\{\beta_{i} \}_{i=0}^{3}$.
	Our Inv-EnNet consists of $K$ coupling layers as depicted in Figure~\ref{fig:framework}(c).
	We empirically average the sub-images to two splits and define $\mathbf{x}_{a}^{0} \!=\! [\alpha_{0}; \alpha_{1}], \mathbf{x}_{b}^{0} \!=\! [\alpha_{2}; \alpha_{3}]$, and $\mathbf{y}_{a}^{K} \!=\! [\beta_{0}; \beta_{1}], \mathbf{y}_{b}^{K} \!=\! [\beta_{2}; \beta_{3}]$ as the input of the invertible network for its forward and backward process respectively. Such of setting is also demonstrated in experiments.
	For the $k^{th}$ layer, its input in the forward process can be represented with $(\mathbf{x}_{a}^{k-1}, \mathbf{x}_{b}^{k-1})$, and the transformation is expressed as:
	\begin{eqnarray}
		\label{eq:cl_f}
		\begin{aligned}
			\mathbf{x}_{b}^{k} &= \mathbf{s}^k_a \circ \mathbf{x}_{b}^{k-1} + \mathbf{t}^k_a \\
			\mathbf{x}_{a}^{k} &= \mathbf{s}^k_b \circ \mathbf{x}_{a}^{k-1} + \mathbf{t}^k_b
		\end{aligned}
	\end{eqnarray}
	with $\mathbf{s}^k_a \!\!=\!\! \exp(h^{k}_a(\mathbf{x}_{a}^{k\!-\!1}))$, $\mathbf{t}^k_a \!\!=\!\! g^{k}_a(\mathbf{x}_{a}^{k\!-\!1})$, $\mathbf{s}^k_b \!\!=\!\! \exp(h^{k}_b(\mathbf{x}_{b}^{k}))$, and $\mathbf{t}^k_b \!\!=\!\! g^{k}_b(\mathbf{x}_{b}^{k})$.
	Note that the $\mathbf{s}^k_a$ and $\mathbf{t}^k_a$ apply the same transformation on the two sub-images of $\mathbf{x}_{b}^{k-1}$.
	In particular, $h^k_a(\cdot)$, $h^k_b(\cdot)$, $g^k_a(\cdot)$ and $g^k_b(\cdot)$ are non-linear mappings for scale and translation respectively, which are modeled by light-weight CNNs and their weights are shared in forward and backward.
	After $K$ coupling layers, we then achieve the outputs of the enhanced sub-images $\{\hat{\alpha}_{i}\}_{i=0}^{3}$ with $\hat{\mathbf{x}}_{a}^{K}\!\!=\!\![\hat{\alpha}_{0};\hat{\alpha}_{1}]$ and $\hat{\mathbf{x}}_{b}^{K}\!\!=\!\![\hat{\alpha}_{2};\hat{\alpha}_{3}]$.
	In the backward process, taking $(\mathbf{y}_{a}^{k}, \mathbf{y}_{b}^{k})$ as input, the $k^{th}$ layer of reverse operation can be performed as:
	\begin{eqnarray}
		\label{eq:cl_b}
		\begin{aligned}
			\mathbf{y}_{a}^{k-1} &= \left( \mathbf{y}_{a}^{k} - (\mathbf{t}^k_b)' \right)\ /\ (\mathbf{s}^k_b)' \\
			\mathbf{y}_{b}^{k-1} &= \left( \mathbf{y}_{b}^{k} - (\mathbf{t}^k_a)' \right) \  /\  (\mathbf{s}^k_a)'
		\end{aligned}
	\end{eqnarray}
	with $(\mathbf{s}^k_a)' \!\!=\!\! \exp\left(h^{k}_a(\mathbf{y}_{a}^{k-1})\right)$, $(\mathbf{t}^k_a)' \!\!=\!\! g^{k}_a(\mathbf{y}_{a}^{k-1})$, $(\mathbf{s}^k_b)' \!\!=\!\! \exp\left(h^{k}_b(\mathbf{y}_{b}^{k})\right)$, and $(\mathbf{t}^k_b)' \!\!=\!\! g^{k}_b(\mathbf{y}_{b}^{k})$.
	Passing the $K$ coupling layers reversely, the degraded low-light sub-images then can be generated and represented as $\{\hat{\beta}_{i}\}_{i=0}^{3}$ with $\hat{\mathbf{y}}_{a}^{0}\!=\![\hat{\beta}_{0};\hat{\beta}_{1}]$ and $\hat{\mathbf{y}}_{b}^{0}\!=\![\hat{\beta}_{2};\hat{\beta}_{3}]$.
	After that, the synthesized paired datasets $\{(\alpha_{i}, \hat{\beta}_{i})\}_{i=0}^{3}$ and $\{(\hat{\alpha}_{i}, \beta_{i})\}_{i=0}^{3}$ are further leveraged to train different discriminators for adversarial learning.

	\subsection{Progressive Self-guided Enhancement}
	\label{sec:2.2}
	Squeezing operation provides a solution to split the image into two parts for the affine coupling layer. However, shown as in Figure~\ref{fig:framework}(c), when we compose the enhanced sub-images directly according to the spatial checkerboard pattern by using pixelshuffle, it would produce severe checkerboard artifacts in the final output.
	To address this issue, we propose a progressive self-guided manner to enhance the image.
	Given the test image $\mathbf{I}$, we aim to enhance it progressive by using the intermediate translation coefficients, i.e., $\mathbf{s}^k_a$ and $\mathbf{t}^k_a$ obtained in forward process, as follows:
	\begin{eqnarray}
		\label{test_phase}
		\mathbf{I}_{k+1} = \upsilon(\mathbf{s}^k_a) \circ \mathbf{I}_{k} + \upsilon(\mathbf{t}^k_a)
	\end{eqnarray}
	where $\upsilon$ denotes the 2$\times$ up-sampling.
	Here, we have $\mathbf{I}_{0} = \mathbf{I}$ as the input and $\mathbf{I}_{K} \!=\! \hat{\mathbf{I}}$ as the final enhancement result.

	\subsection{Learning Objective}
	\label{sec:2.3}
	To make the training stable and produce visual pleasing results, we design various loss functions for the proposed Inv-EnNet.

	\noindent \textbf{Adversarial Loss.}
	We use two PatchGAN discriminators for adversarial learning. Typically, the LSGAN~\cite{arxiv1611.04076} loss is adopted and for the forward and backward process, we have:
	\begin{eqnarray}
		\label{eq:gan_f}
		\begin{aligned}
			\mathcal{L}_{ADV}^{f} \!\! & =\! \min_G \max_{D^N} \mathbb{E}_{\beta}[D^N(\beta)] \!+\! \mathbb{E}_{{\alpha}}[(D^N(G({\alpha}))\!-\!1)^2] \\
			\mathcal{L}_{ADV}^{b} \!\! & =\! \min_{G^{-1}} \max_{D^L} \mathbb{E}_{\alpha}[D^L(\alpha)] \!+\! \mathbb{E}_{{\beta}}[(D^L(G^{-1}({\beta}))\!-\!1)^2]
		\end{aligned}
	\end{eqnarray}
	where $G$ denotes the forward process of Inv-EnNet and $G^{-1}$ corresponds to the backward process.

	\noindent \textbf{Transformation-Consistent Loss.}
	From Figure~\ref{fig:framework}(c), the affine coupling layer takes two parts as input and provides guidance to each other in an alternating manner. To improve the stability of the proposed Inv-EnNet, we claim that the translations for enhancement in each coupling layer should be as similar as possible.
	Therefore, we introduce the transformation-consistent loss both in forward and backward processes to constrain the distances between the alternating translations as follows:
	\begin{eqnarray}
		\label{eq:si_f}
		\begin{aligned}
			\mathcal{L}^{f}_{TC} & = \sum\nolimits_k \left \| \mathbf{s}^k_a - \mathbf{s}^k_b \right \|_1 + \left \| \mathbf{t}^k_a - \mathbf{t}^k_b \right \|_1 \\
			\mathcal{L}^{b}_{TC} & = \sum\nolimits_k \left \| (\mathbf{s}^k_a)' - (\mathbf{s}^k_b)' \right \|_1 + \left \| (\mathbf{t}^k_a)' - (\mathbf{t}^k_b)' \right \|_1
		\end{aligned}
	\end{eqnarray}
	The effect of $\mathcal{L}_{TC}$ on model performance would be further demonstrated in the ablation study.

	\noindent \textbf{Detail-Preserving Loss.}
	In order to preserve the image details, EnlightenGAN~\cite{jiang2021enlightengan} constrains the distance of VGG-features between the low light input and the model output.
	However, the inconsistent illuminance between a low light and a normal light image may affect the extraction of VGG-features, and lead to performance degradation.
	Therefore, we introduce the guided filter~\cite{10.5555/1886063.1886065}, i.e., $\mathcal{G}$, to preserve details from low-light image but share the same illumination with the network output.
	Then the loss is defined as:
	\begin{eqnarray}
		\label{eq:dp_f}
		\begin{aligned}
			\mathcal{L}^{f}_{DP} & = \sum\nolimits_i \left \| \phi_i(\alpha^\mathcal{G})-\phi_i(\hat{\alpha}) \right \|_2 \\
			\mathcal{L}^{b}_{DP} & = \sum\nolimits_i \left \| \phi_i(\beta^\mathcal{G})-\phi_i(\hat{\beta}) \right \|_2
		\end{aligned}
	\end{eqnarray}
	where $\alpha^\mathcal{G} = \mathcal{G}(\alpha, \hat{\alpha})$ and $\beta^\mathcal{G} = \mathcal{G}(\beta, \hat{\beta})$, $\phi_i(\cdot)$ refers to the extraction of VGG-features from the $i^{th}$ max pooling layer of VGG-16 model pre-trained on ImageNet.

	\noindent \textbf{Reversibility Loss.}
	The enhanced and degraded results might contain some pixels with values exceeding the range of [0,1], thereby leading to over-exposed area and losing information of low light images.

	Fortunately, the merits of the invertible architecture provides us a solution to mitigate this issue.
	Inspired by the cycle-consistency loss adopted in the unpaired image translation, we design a reversibility loss with a combination of the $\min$ and $\max$ operations:
	\begin{eqnarray}
		\label{eq:re_f}
		\begin{aligned}
			\mathcal{L}_{R}^{f} & = \sum\nolimits_i \left \| \alpha_i -G^{-1}(\min(\max(\hat{\alpha}_i, 1), 0)) \right \|_1 \\
			\mathcal{L}_{R}^{b} & = \sum\nolimits_i \left \| \beta_i - G(\min(\max(\hat{\beta}_i, 1), 0)) \right \|_1
		\end{aligned}
	\end{eqnarray}
	Finally, the overall loss function of our model is:
	\begin{equation}
		\begin{aligned}
			\mathcal{L} & = (\mathcal{L}_{ADV}^{f} + \mathcal{L}_{ADV}^{b}) + \eta (\mathcal{L}_{TC}^{f} + \mathcal{L}_{TC}^{b}) \\
			& + \lambda (\mathcal{L}_{DP}^{f} + \mathcal{L}_{DP}^{b}) + \mu (\mathcal{L}_{R}^{f}+\mathcal{L}_{R}^{b})
		\end{aligned}
	\end{equation}
	where $\lambda$, $\mu$ and $\eta$ are weights to control the relative importance of the objectives.
	\begin{figure*}[htbp]
		\centering
		\begin{subfigure}[b]{0.19\textwidth}
			\centering
			\includegraphics[width=\textwidth]{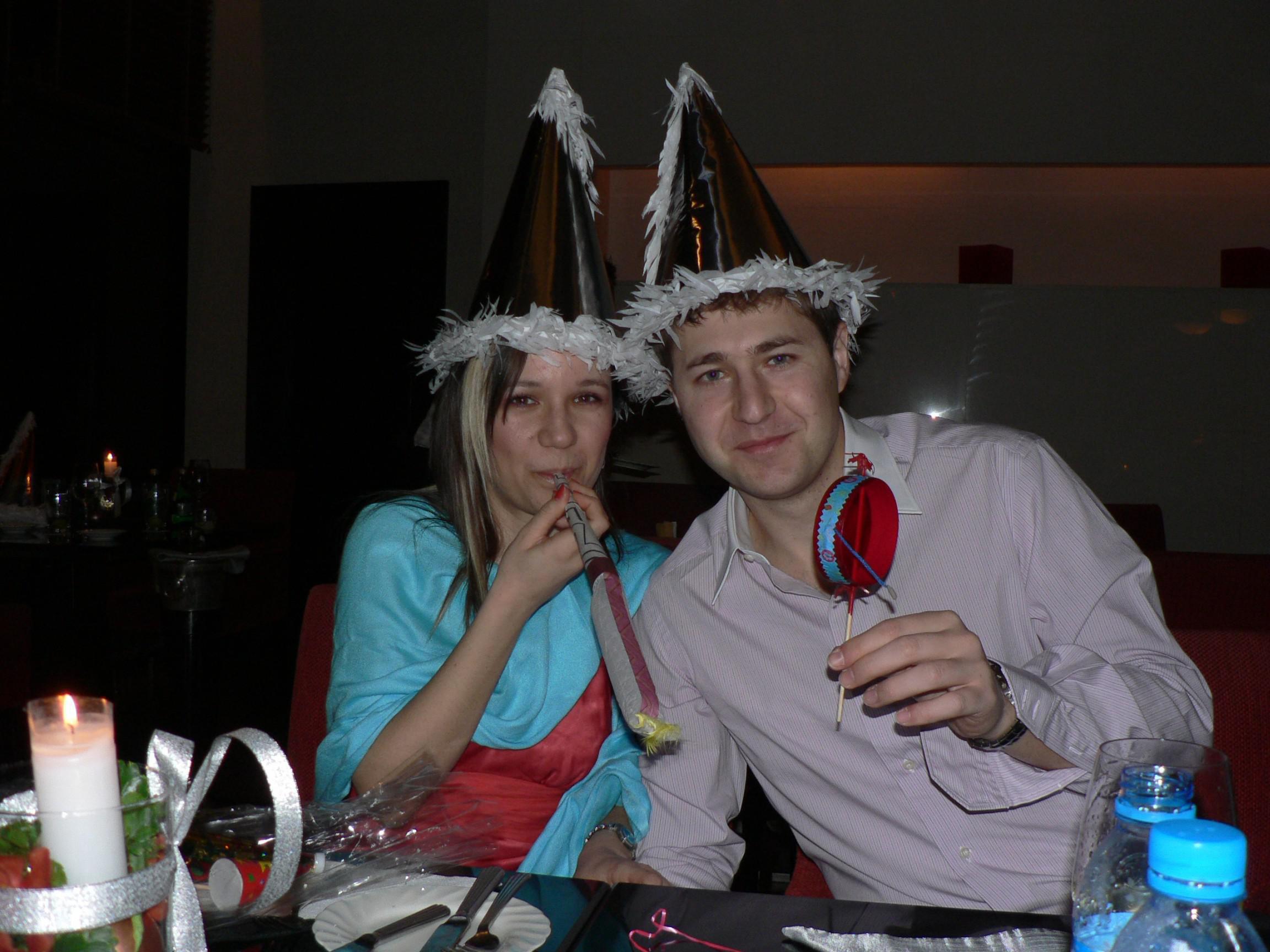}
			\caption{Input}
			\label{qualitative_input}
		\end{subfigure}
		\begin{subfigure}[b]{0.19\textwidth}
			\centering
			\includegraphics[width=\textwidth]{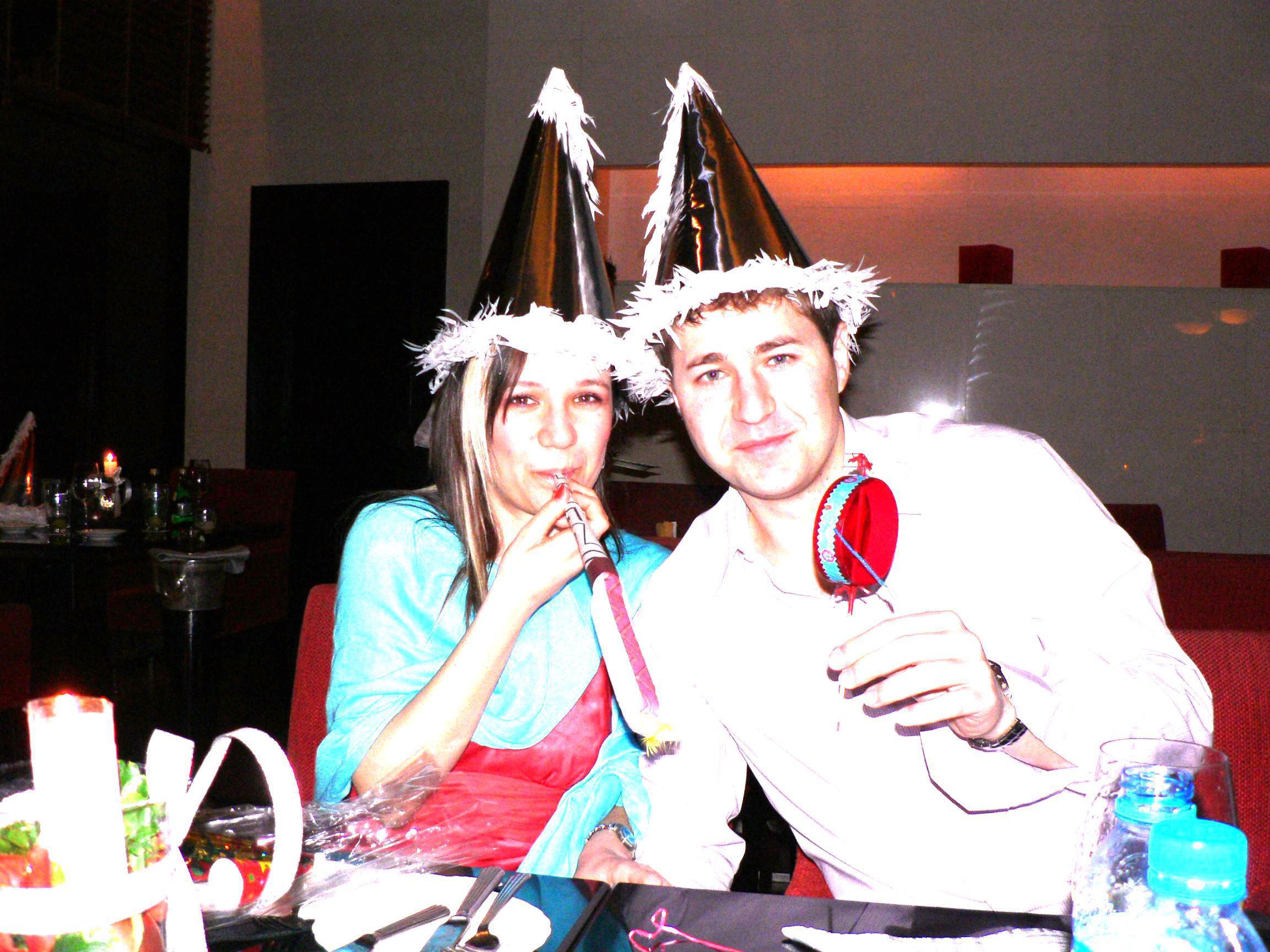}
			\caption{RUAS~\cite{liu2021ruas}}
			\label{qualitative_RUAS}
		\end{subfigure}
		\begin{subfigure}[b]{0.19\textwidth}
			\centering
			\includegraphics[width=\textwidth]{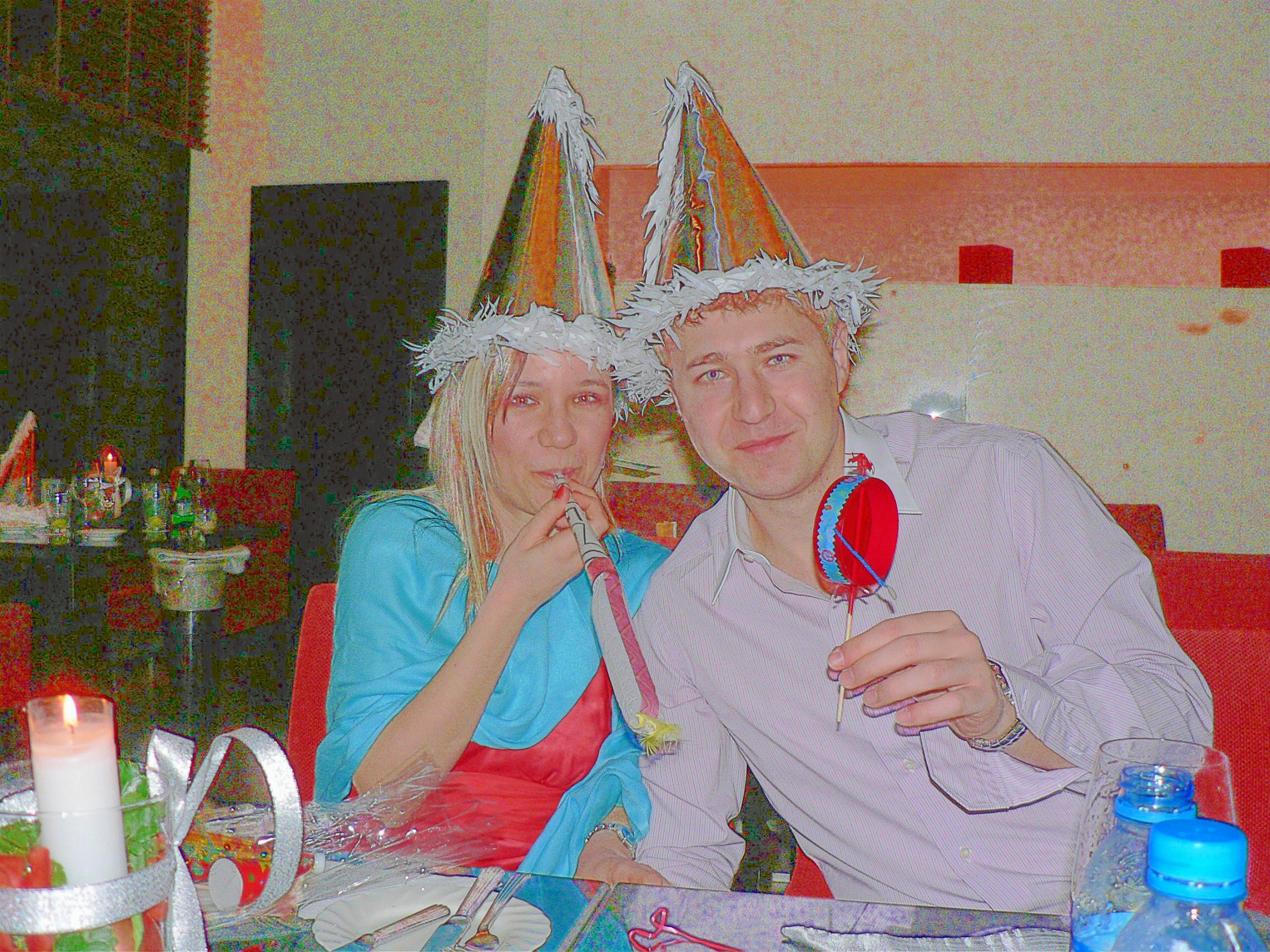}
			\caption{RetinexNet~\cite{Chen2018Retinex}}
			\label{qualitative_RetinexNet}
		\end{subfigure}
		\begin{subfigure}[b]{0.19\textwidth}
			\centering
			\includegraphics[width=\textwidth]{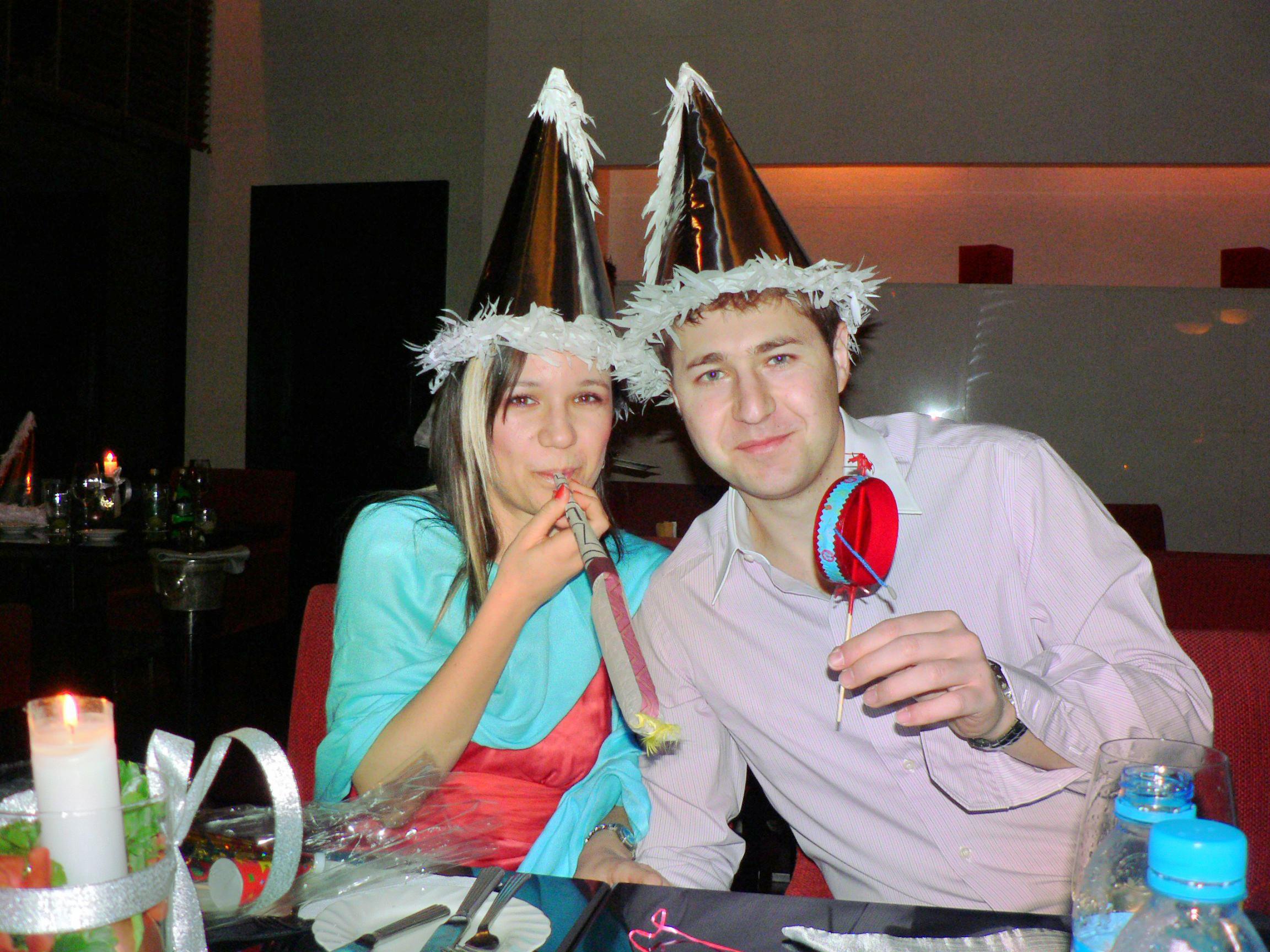}
			\caption{DeepUPE~\cite{Wang_2019_CVPR}}
			\label{qualitative_DeepUPE}
		\end{subfigure}
		\begin{subfigure}[b]{0.19\textwidth}
			\centering
			\includegraphics[width=\textwidth]{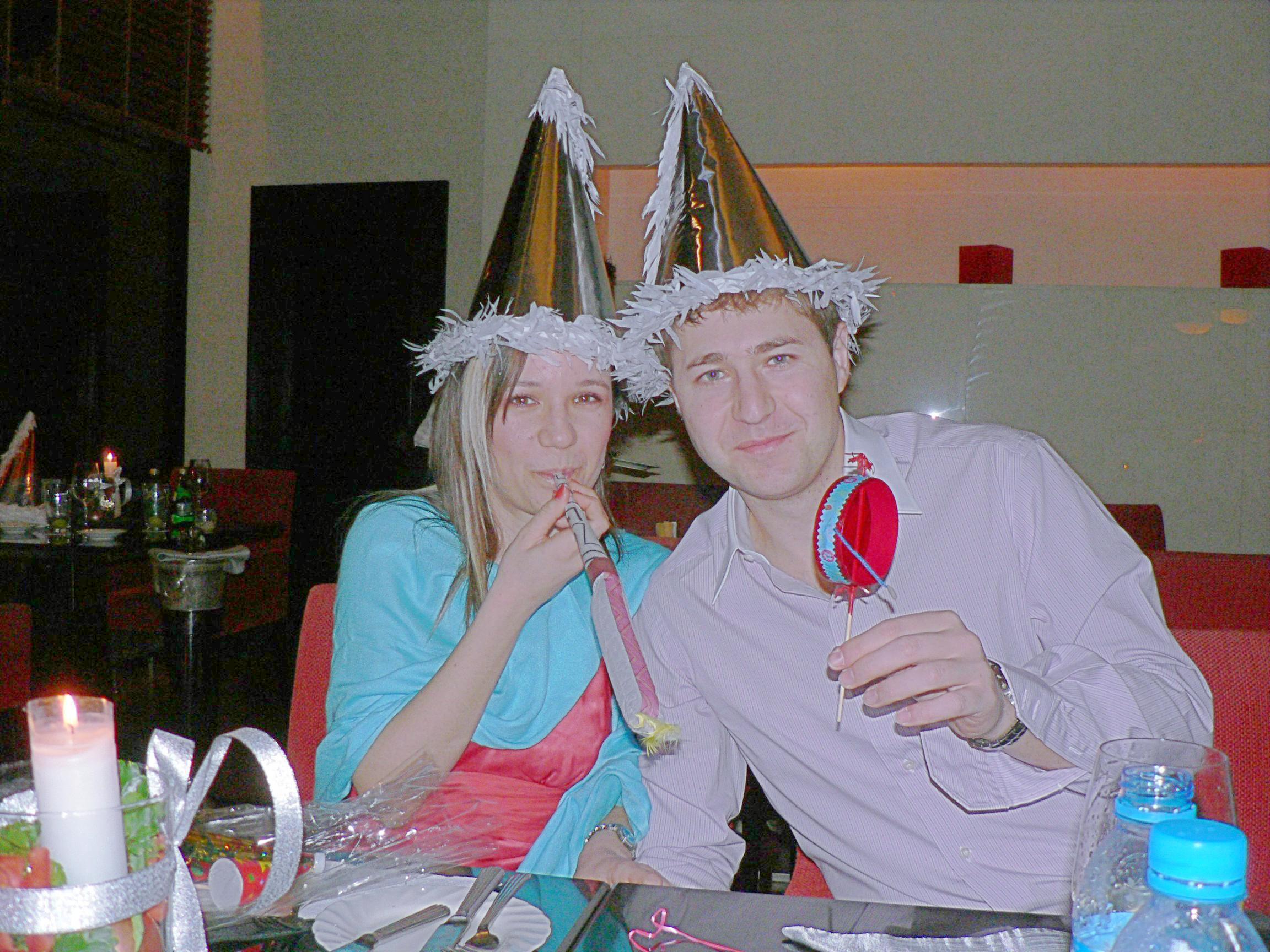}
			\caption{ZeroDCE~\cite{Zero-DCE}}
			\label{qualitative_ZeroDCE}
		\end{subfigure}
		\begin{subfigure}[b]{0.19\textwidth}
			\centering
			\includegraphics[width=\textwidth]{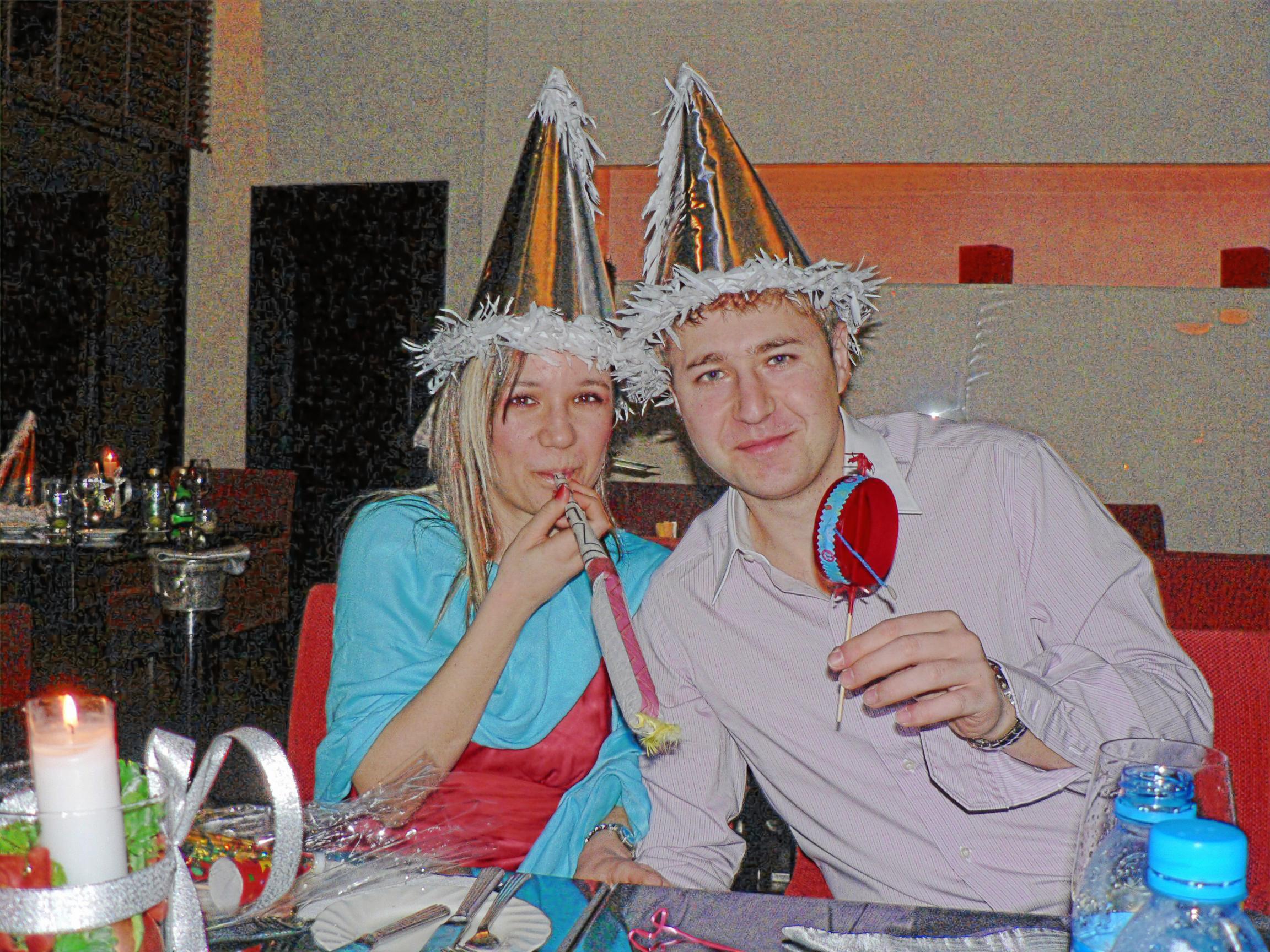}
			\caption{KinD++~\cite{kind++}}
			\label{qualitative_KinD++}
		\end{subfigure}
		\begin{subfigure}[b]{0.19\textwidth}
			\centering
			\includegraphics[width=\textwidth]{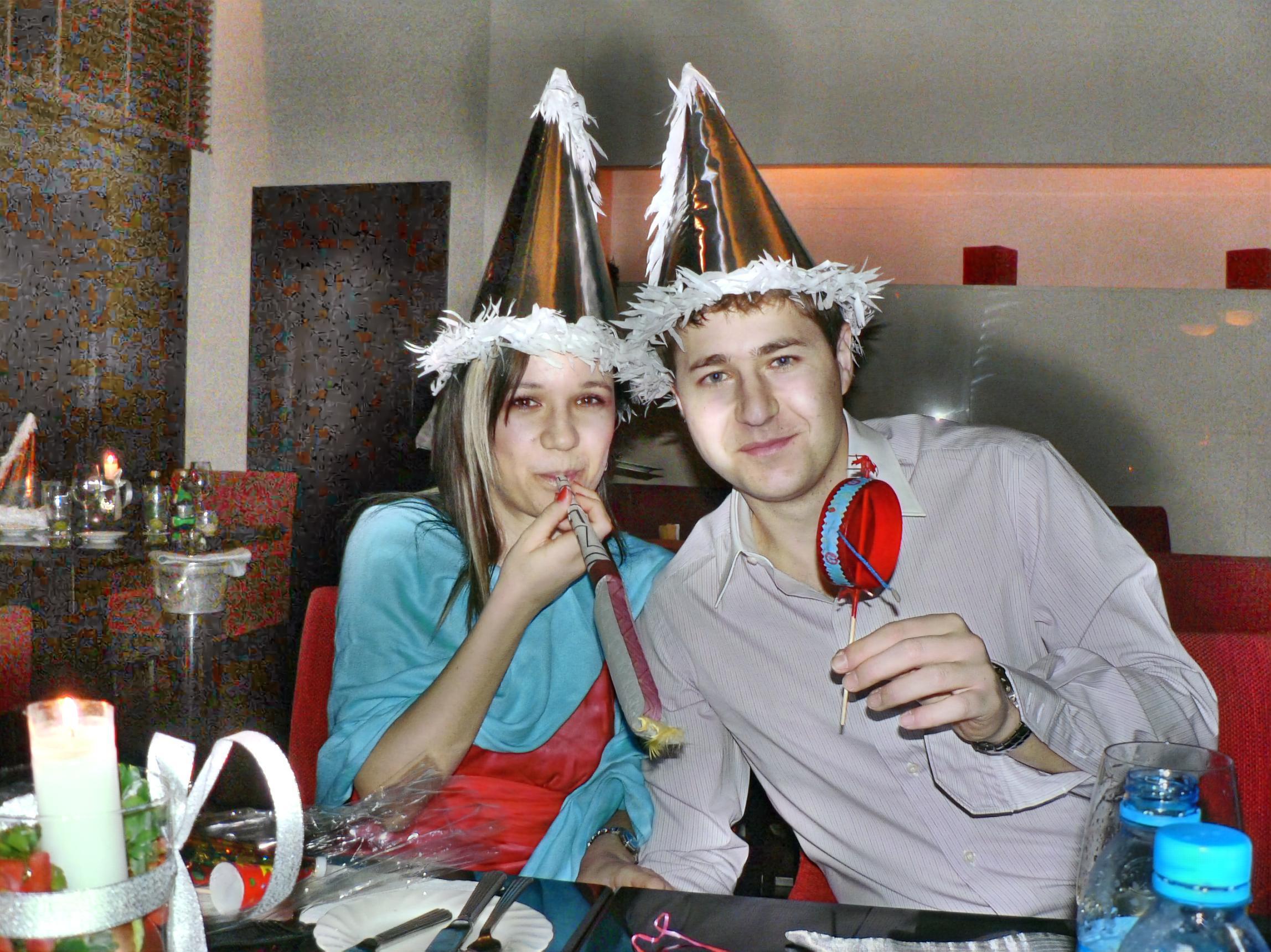}
			\caption{DRBN stage1~\cite{drbn}}
			\label{qualitative_DRBN}
		\end{subfigure}
		\begin{subfigure}[b]{0.19\textwidth}
			\centering
			\includegraphics[width=\textwidth]{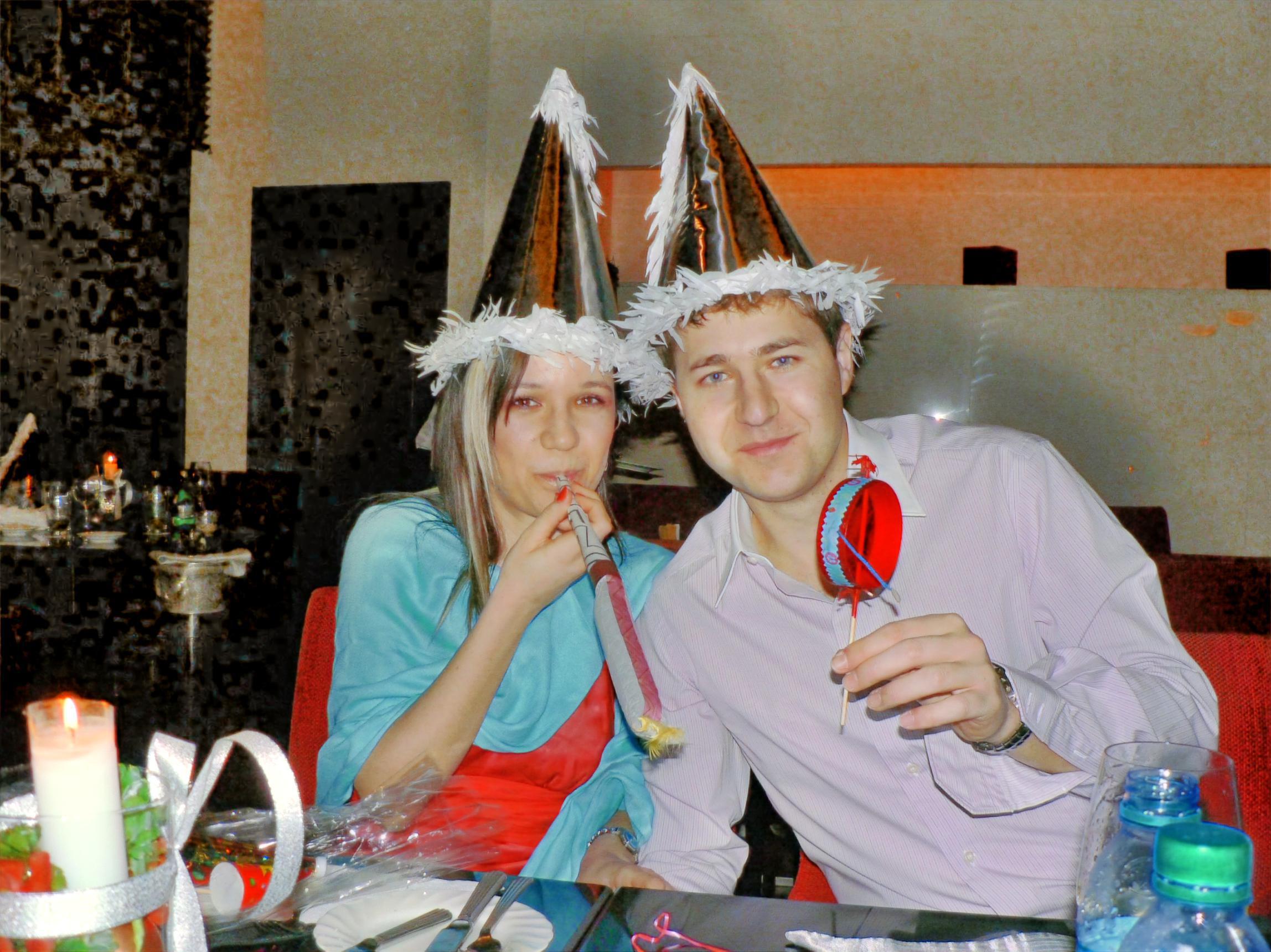}
			\caption{DRBN stage2~\cite{drbn}}
			\label{qualitative_DRBN_2}
		\end{subfigure}
		\begin{subfigure}[b]{0.19\textwidth}
			\centering
			\includegraphics[width=\textwidth]{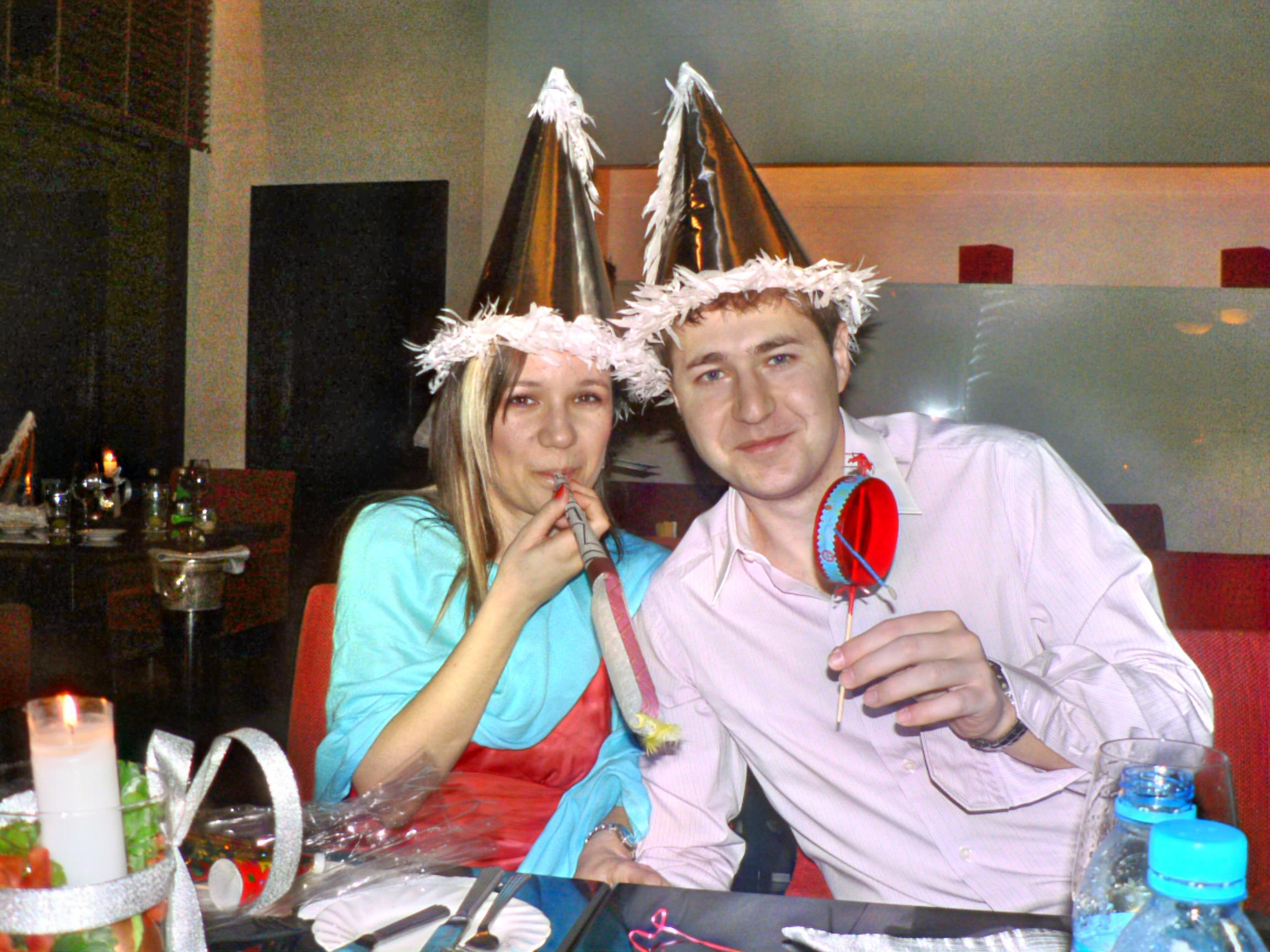}
			\caption{EnlightenGAN~\cite{jiang2021enlightengan}}
			\label{qualitative_EGAN}
		\end{subfigure}
		\begin{subfigure}[b]{0.19\textwidth}
			\centering
			\includegraphics[width=\textwidth]{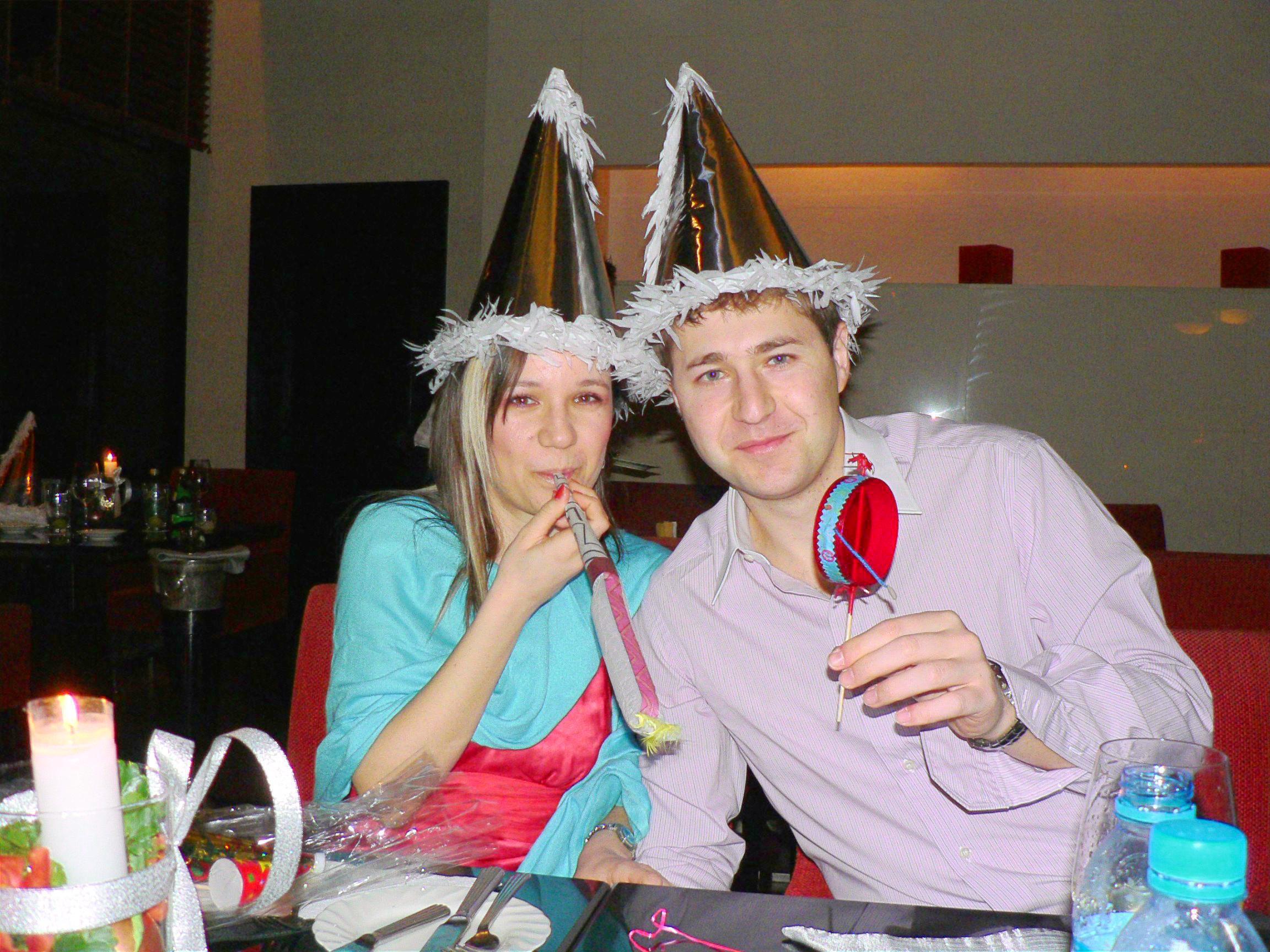}
			\caption{Inv-EnNet (Ours)}
			\label{qualitative_ours}
		\end{subfigure}
		\vspace{-3mm}
		\caption{Visual comparison with SOTA methods on the VV dataset.}
		\label{fig:sota}
		\vspace{-12mm}
	\end{figure*}
	\begin{figure*}[htbp]
		\centering
		\begin{subfigure}[b]{0.238\textwidth}
			\hspace{-1.3mm}
			\includegraphics[width=\textwidth]{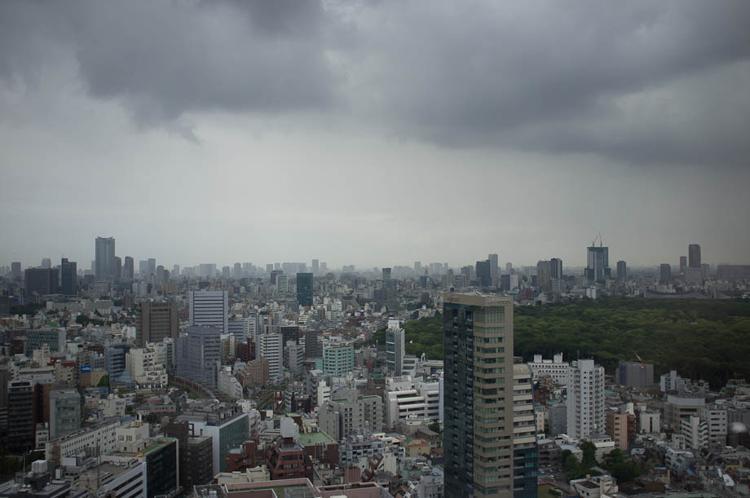}
		\end{subfigure}
		\begin{subfigure}[b]{0.238\textwidth}
			\hspace{-1.3mm}
			\includegraphics[width=\textwidth]{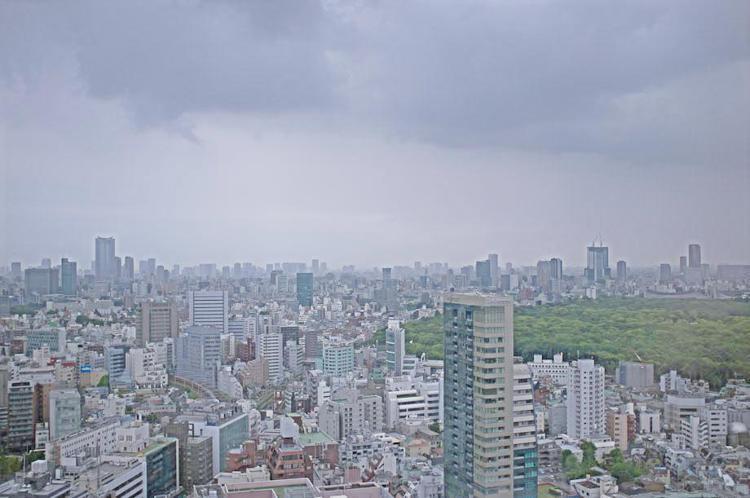}
		\end{subfigure}
		\begin{subfigure}[b]{0.238\textwidth}
			\hspace{-1.3mm}
			\includegraphics[width=\textwidth]{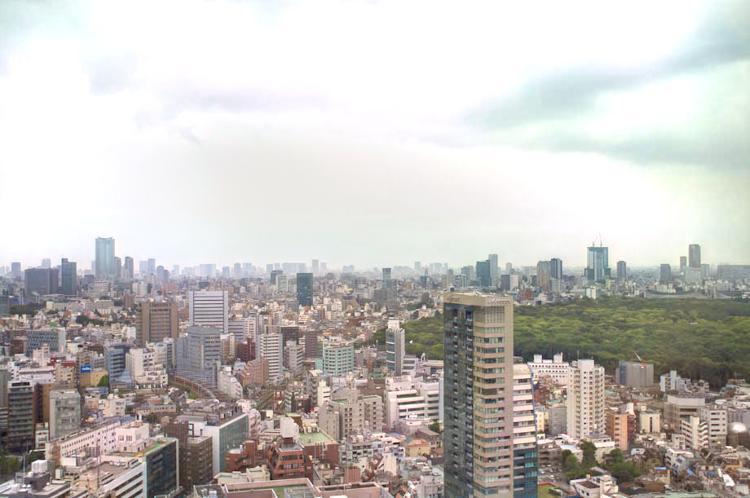}
		\end{subfigure}
		\begin{subfigure}[b]{0.238\textwidth}
			\hspace{-1.3mm}
			\includegraphics[width=\textwidth]{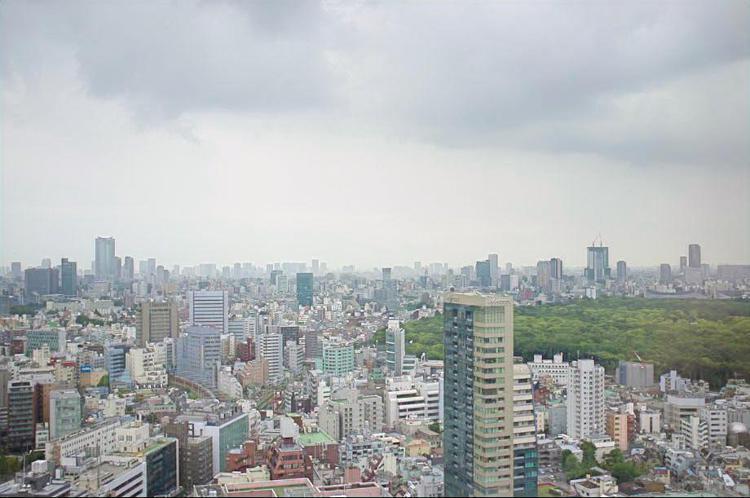}
		\end{subfigure}
		\begin{subfigure}[b]{0.238\textwidth}
			\hspace{-1.3mm}
			\includegraphics[width=\textwidth]{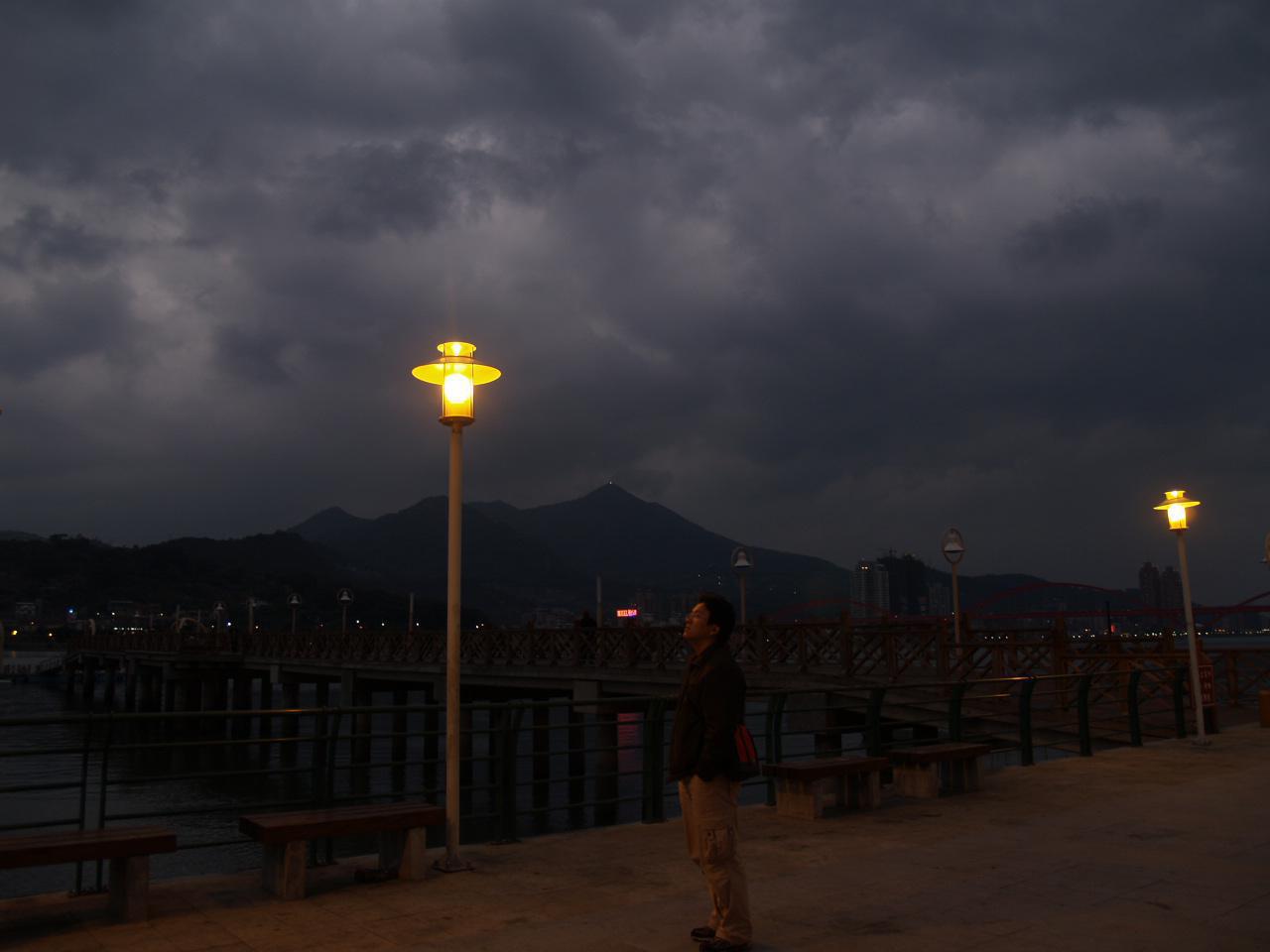}
		\end{subfigure}
		\begin{subfigure}[b]{0.238\textwidth}
			\hspace{-1.3mm}
			\includegraphics[width=\textwidth]{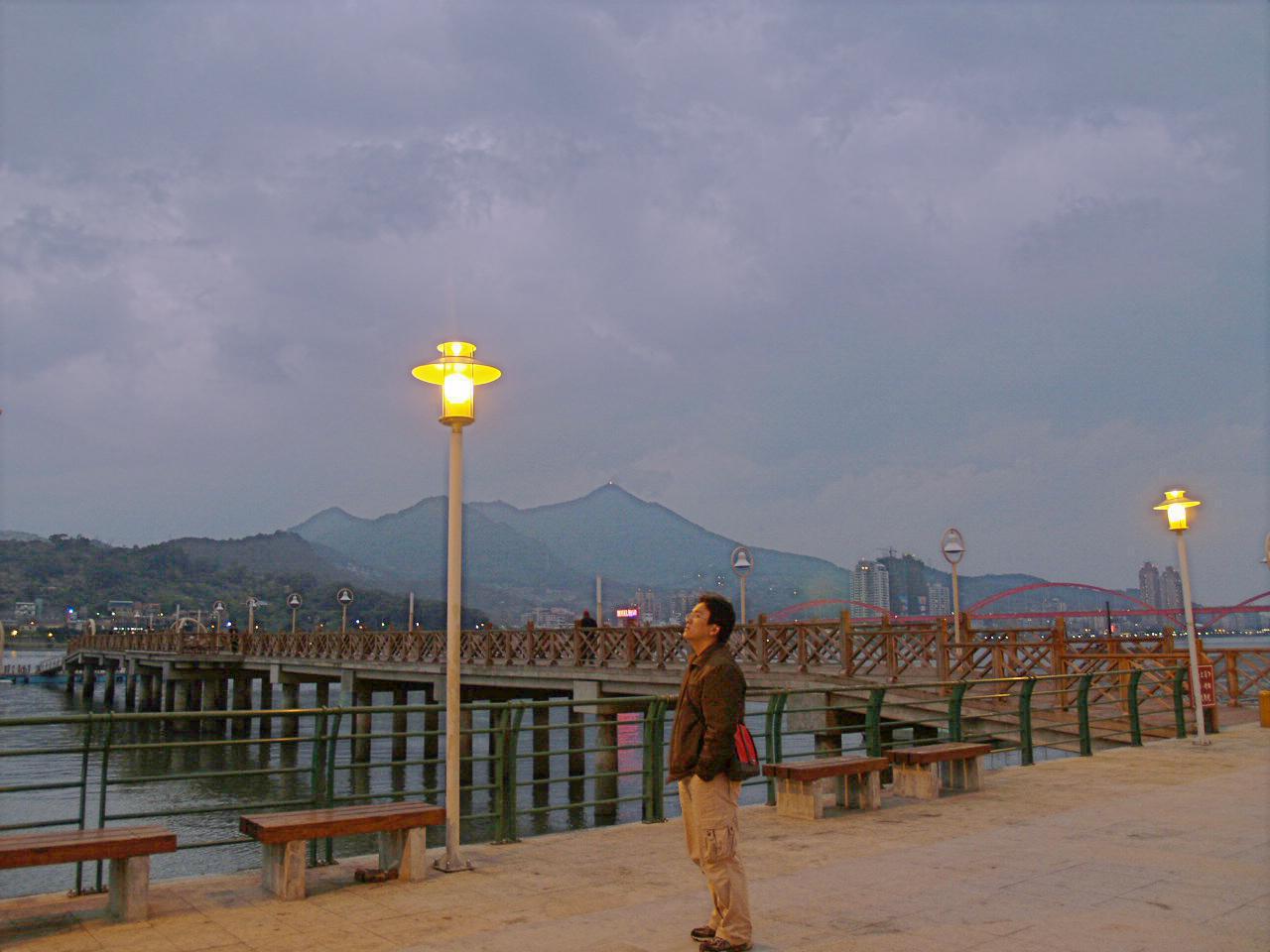}
		\end{subfigure}
		\begin{subfigure}[b]{0.238\textwidth}
			\hspace{-1.3mm}
			\includegraphics[width=\textwidth]{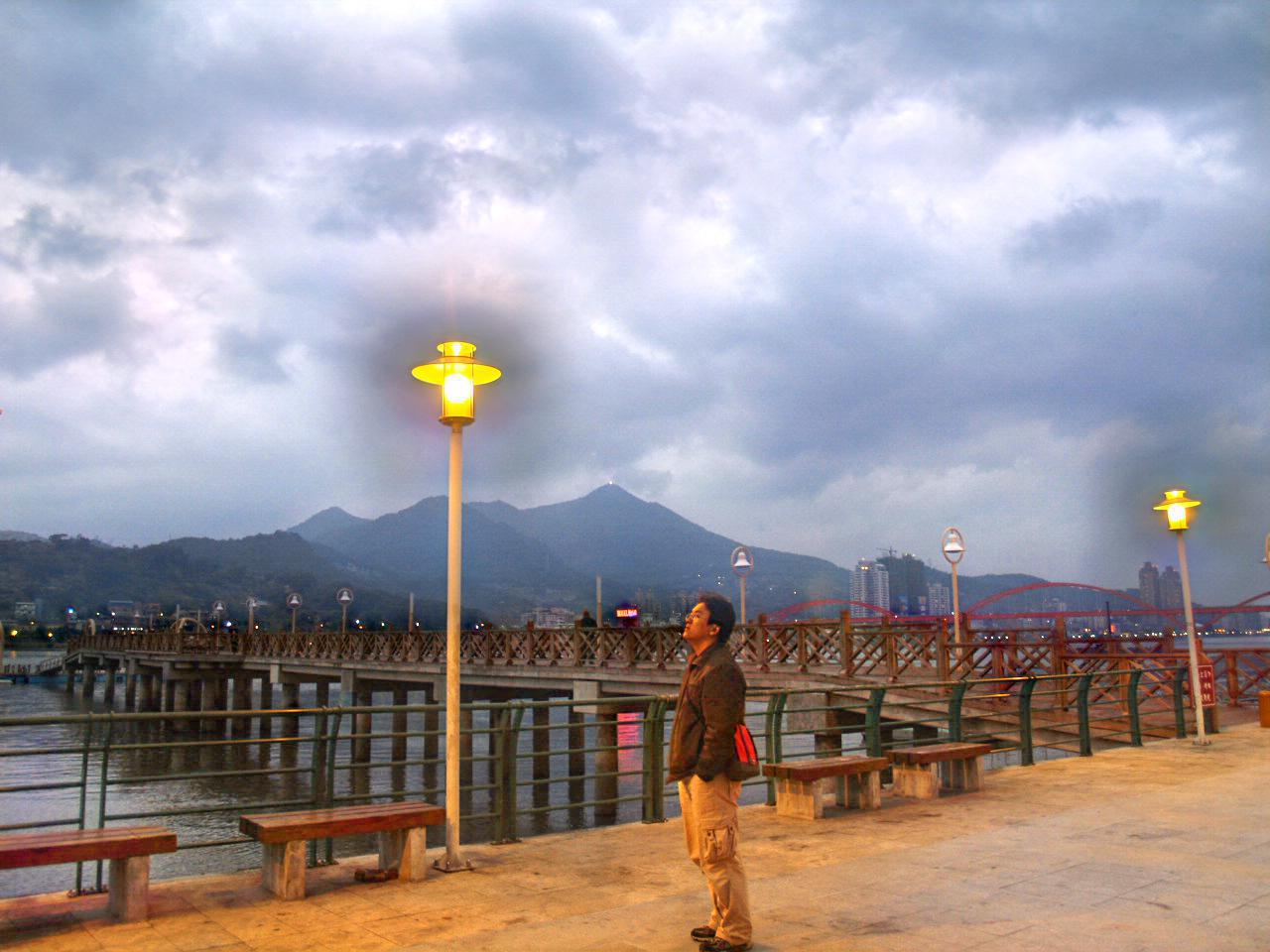}
		\end{subfigure}
		\begin{subfigure}[b]{0.238\textwidth}
			\hspace{-1.3mm}
			\includegraphics[width=\textwidth]{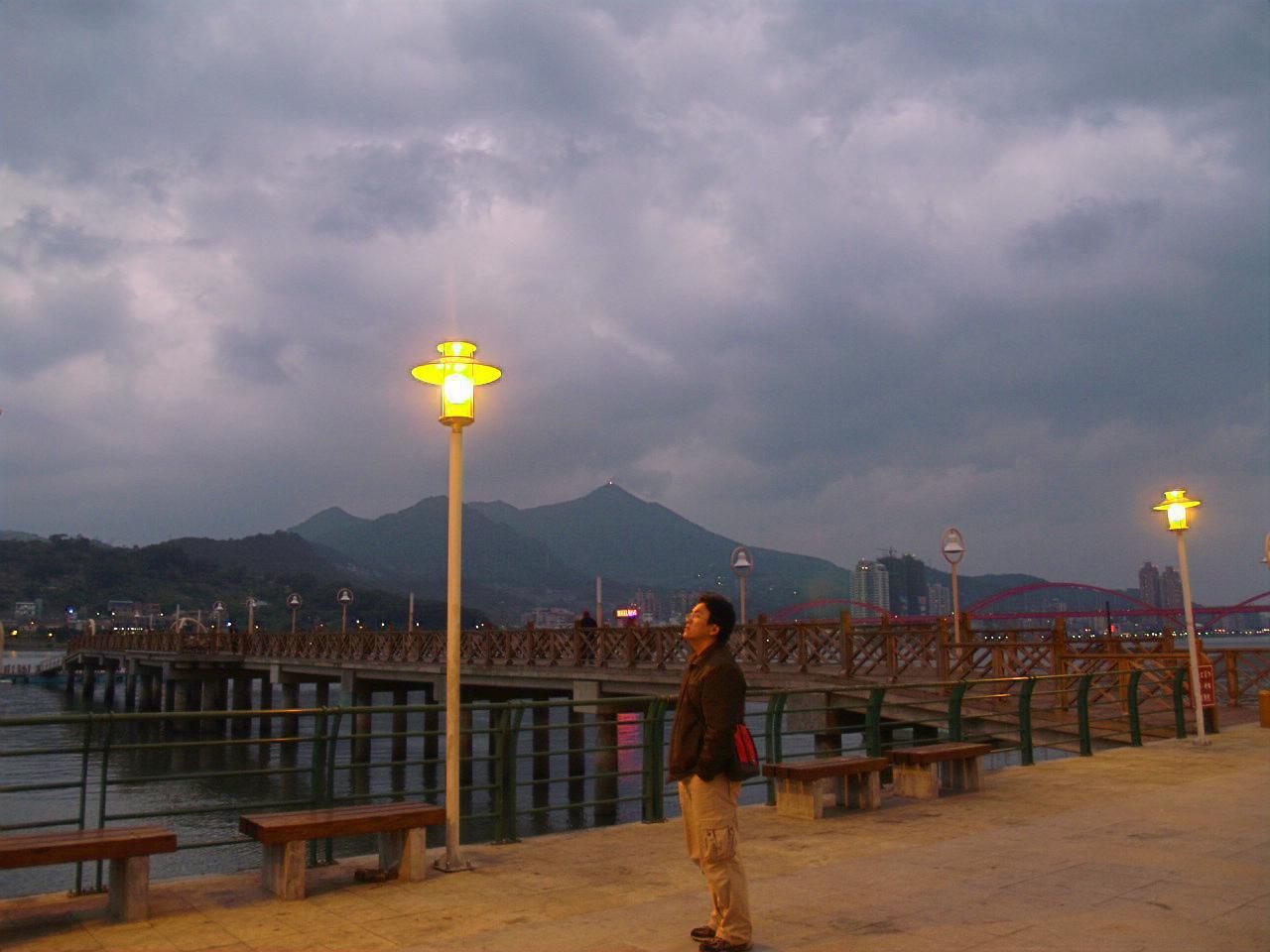}
		\end{subfigure}
		\begin{subfigure}[b]{0.238\textwidth}
			\includegraphics[width=\textwidth]{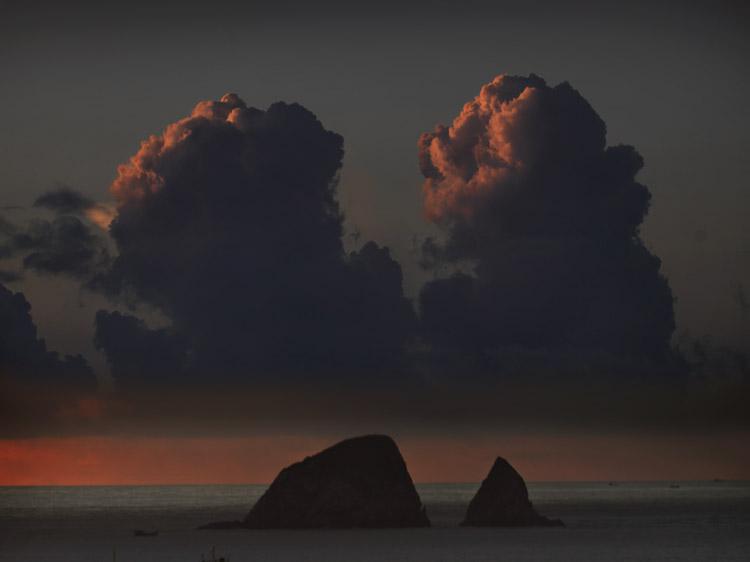}
			\caption{Input}
		\end{subfigure}
		\begin{subfigure}[b]{0.238\textwidth}
			\includegraphics[width=\textwidth]{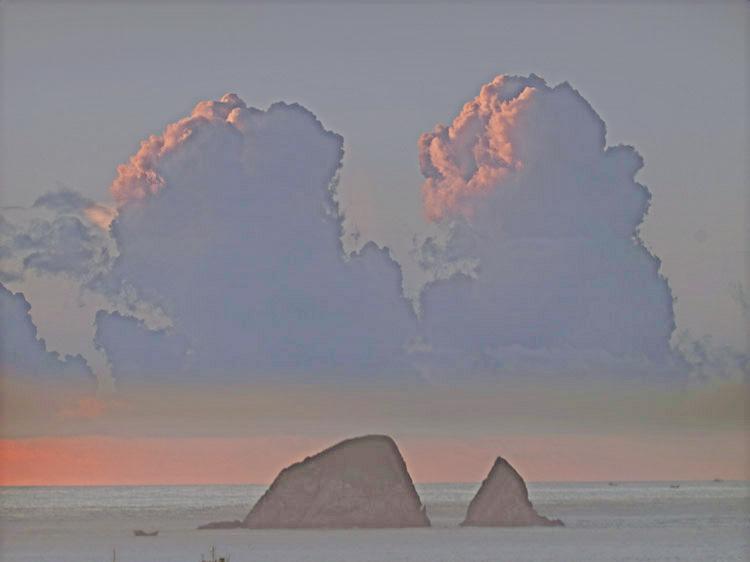}
			\caption{ZeroDCE~\cite{Zero-DCE}}
		\end{subfigure}
		\begin{subfigure}[b]{0.238\textwidth}
			\includegraphics[width=\textwidth]{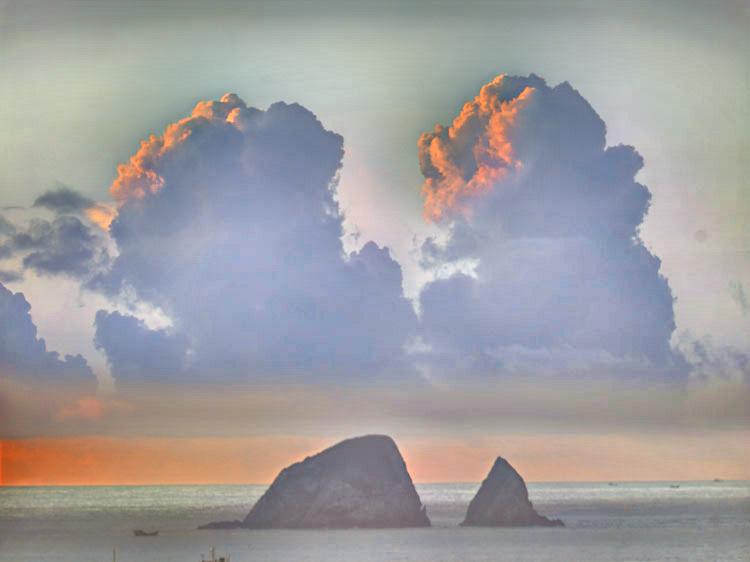}
			\caption{EnlightenGAN~\cite{jiang2021enlightengan}}
		\end{subfigure}
		\begin{subfigure}[b]{0.238\textwidth}
			\includegraphics[width=\textwidth]{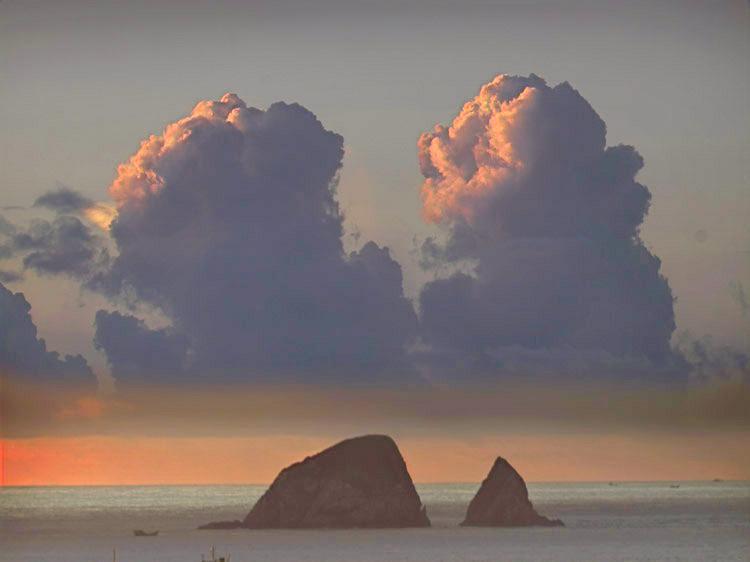}
			\caption{Inv-EnNet (Ours)}
		\end{subfigure}
		\vspace{-3mm}
		\caption{Visual comparison with current unsupervised SOTA methods on several challenging low-light images from NPE.}
		\label{fig:sota_unsupervised}
		\vspace{-12mm}
	\end{figure*}
	\begin{figure*}[htbp]
		\centering
		\begin{subfigure}[t]{0.19\textwidth}
			\centering
			\setlength{\abovecaptionskip}{2.6mm}
			\includegraphics[width=\textwidth]{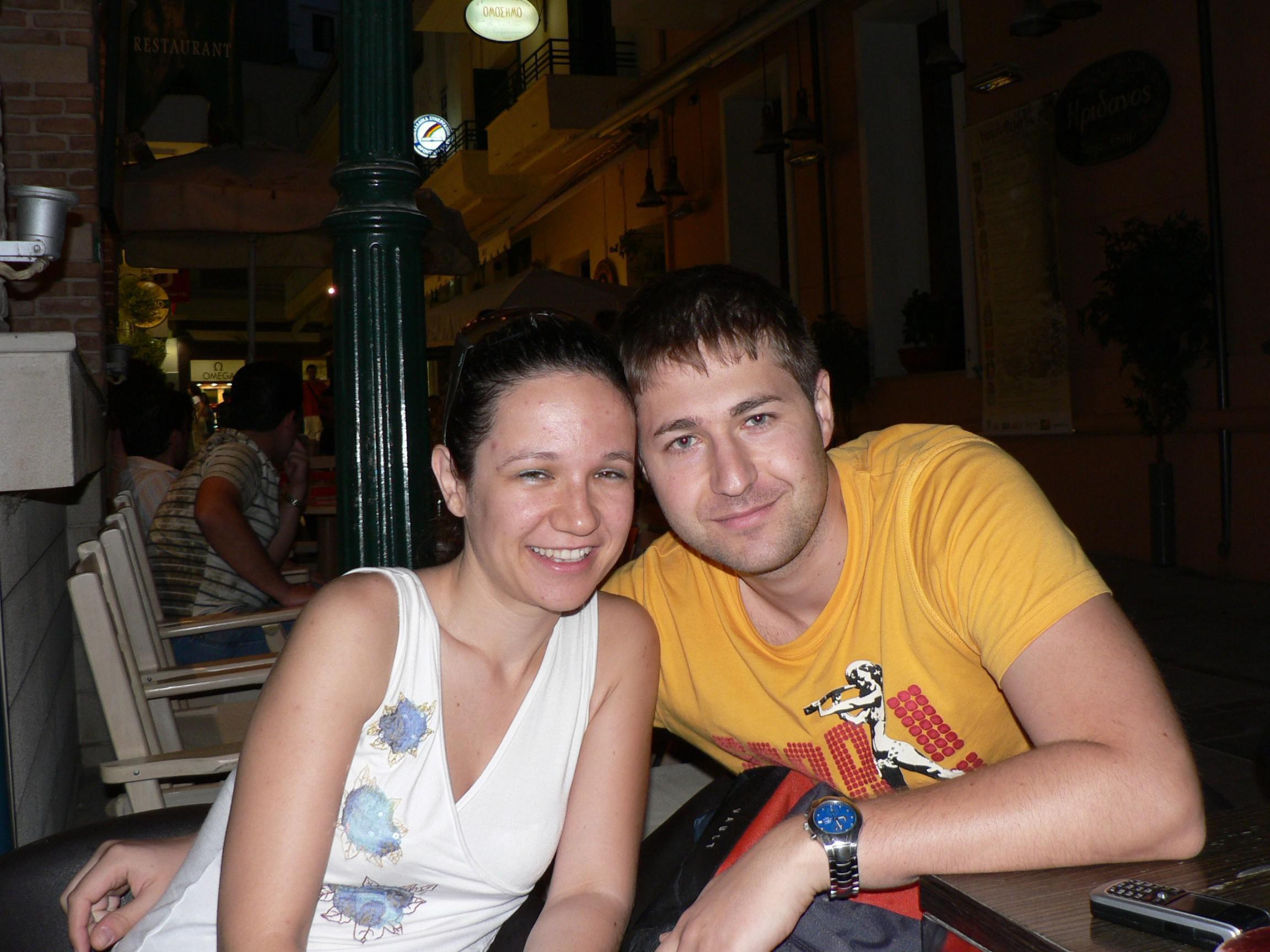}
			\caption{Input}
			\label{main_ablation_in}
		\end{subfigure}
		\begin{subfigure}[t]{0.19\textwidth}
			\centering
			\includegraphics[width=\textwidth]{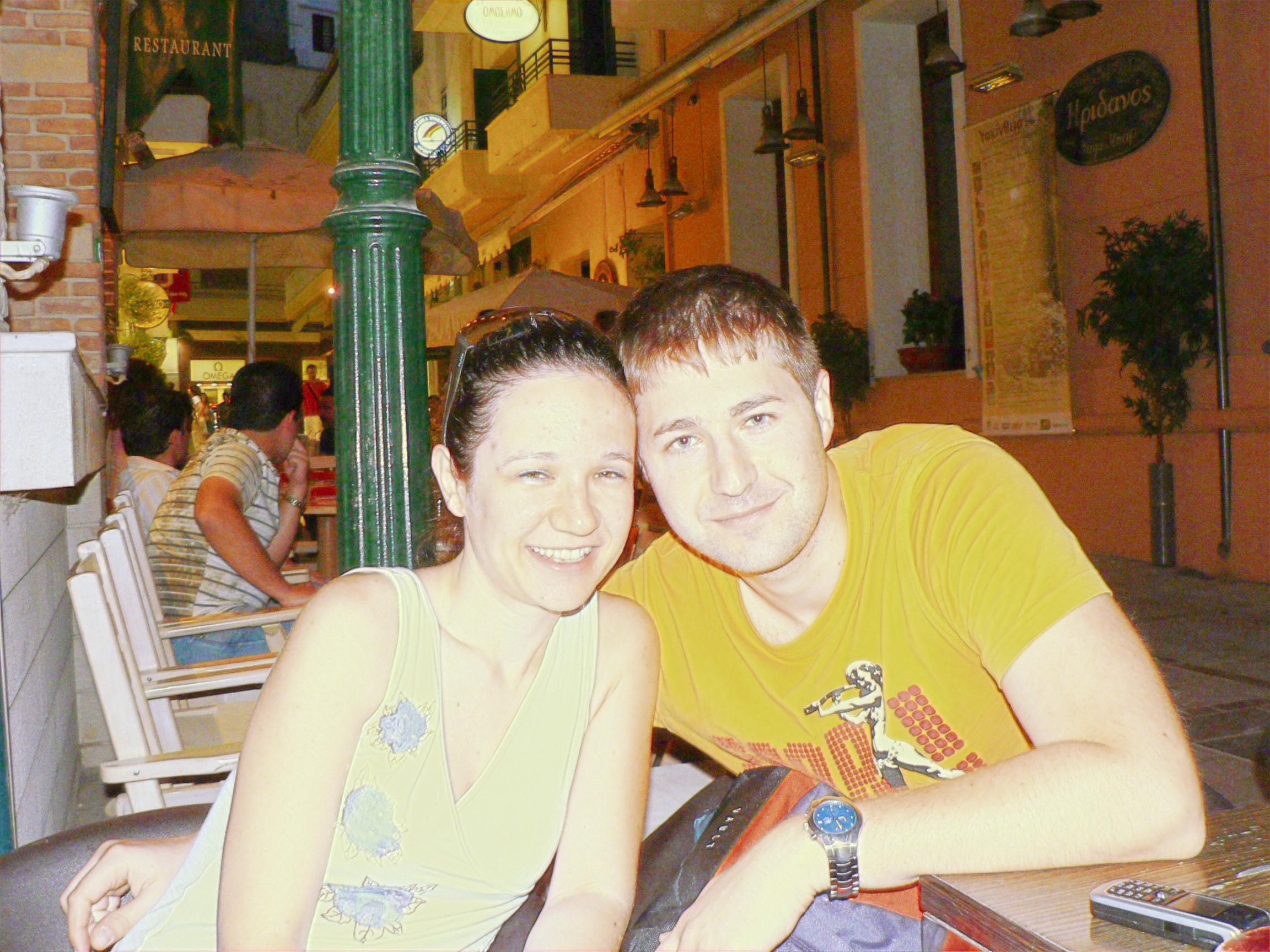}
			\caption{ $\text{Inv-EnNet}^{w/o}_{\mathcal{L}_{TC}}$ }
			\label{main_ablation_no_si}
		\end{subfigure}
		\begin{subfigure}[t]{0.19\textwidth}
			\centering
			\includegraphics[width=\textwidth]{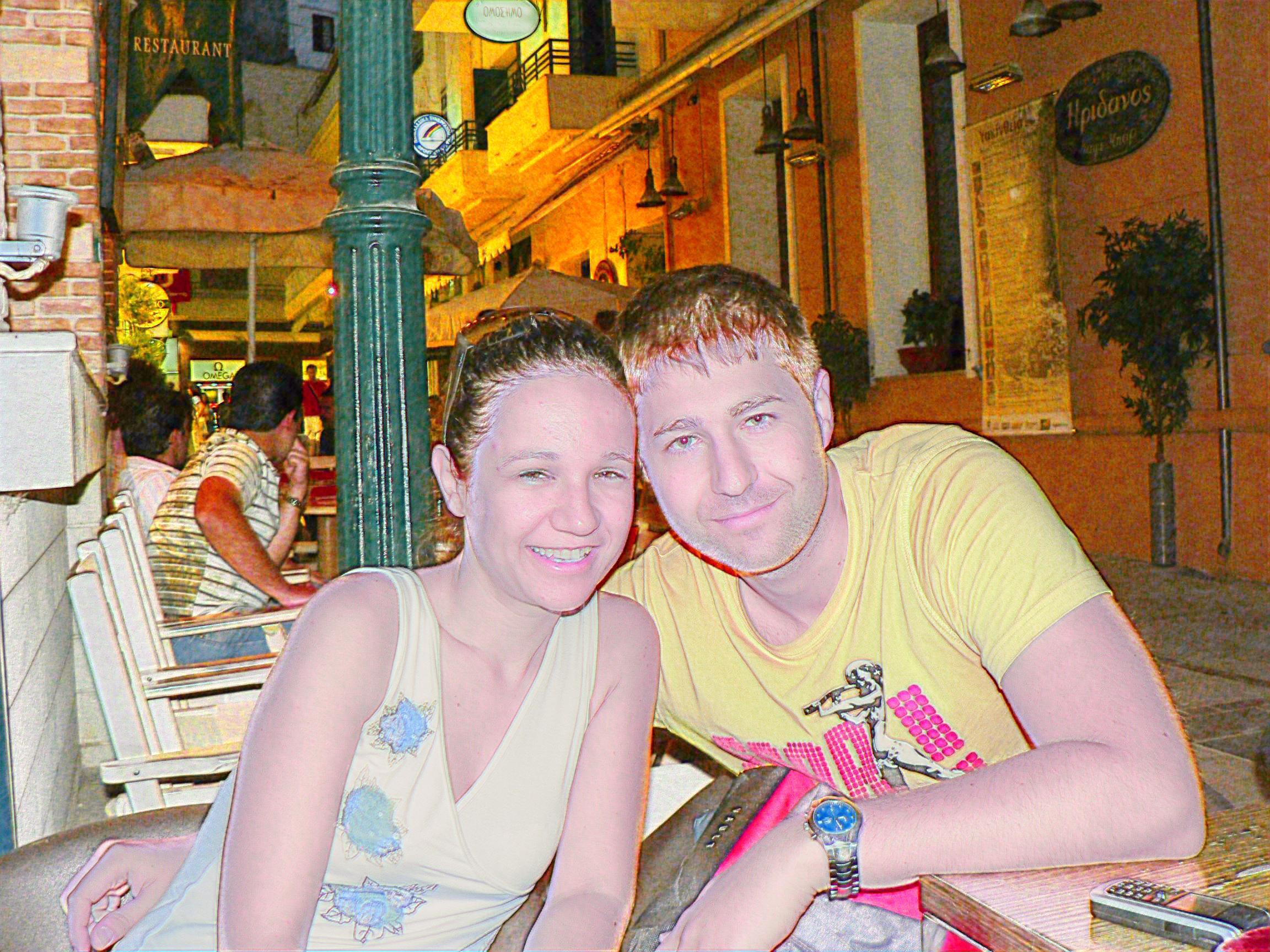}
			\caption{ $\text{Inv-EnNet}^{w/o}_{\mathcal{L}_{DP}}$}
			\label{main_ablation_no_vgg}
		\end{subfigure}
		\begin{subfigure}[t]{0.19\textwidth}
			\centering
			\includegraphics[width=\textwidth]{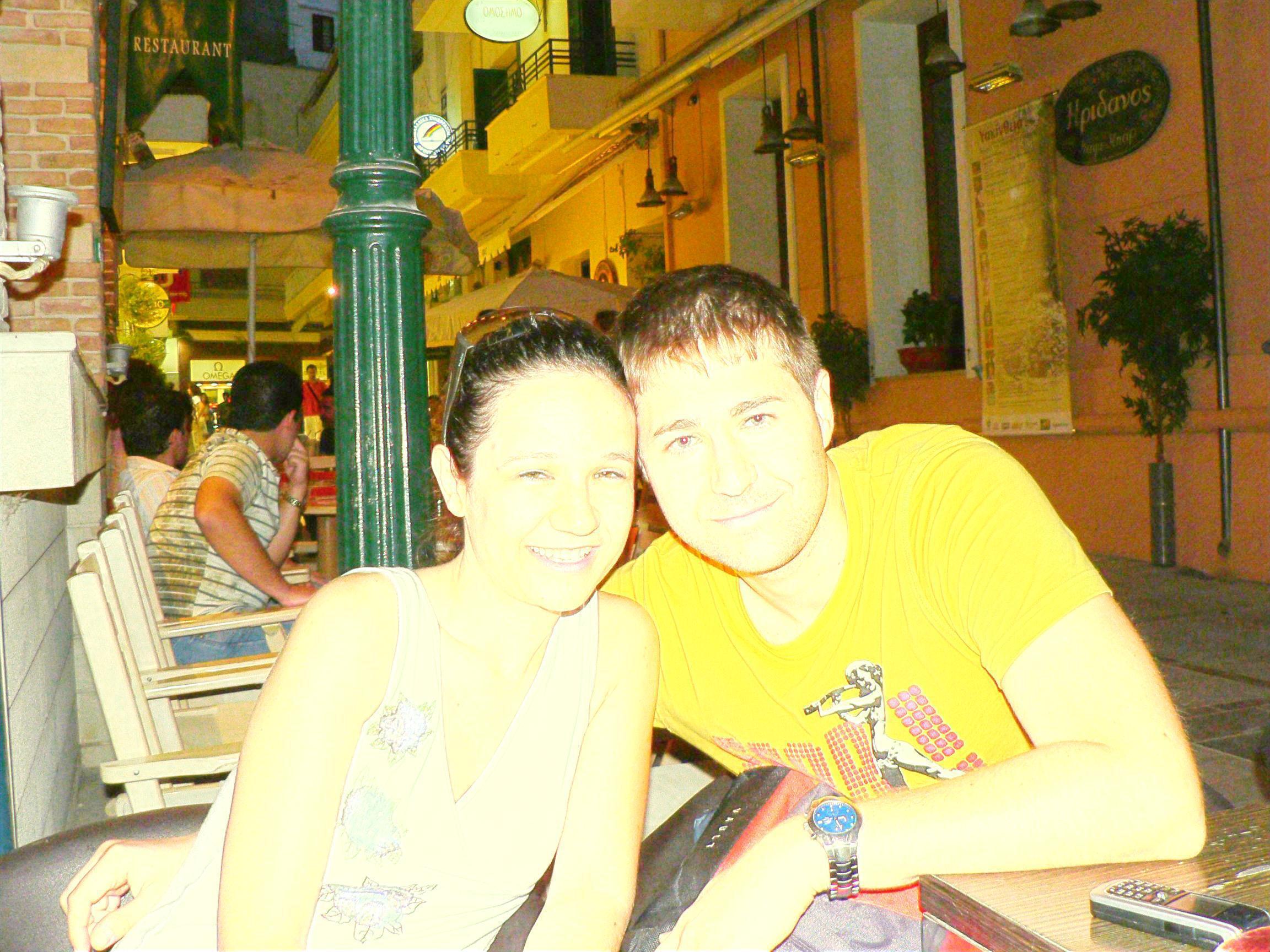}
			\caption{ $\text{Inv-EnNet}^{w/o}_{\mathcal{L}_{R}}$}
			\label{main_ablation_no_cons}
		\end{subfigure}
		\begin{subfigure}[t]{0.19\textwidth}
			\centering
			\setlength{\abovecaptionskip}{2.6mm}
			\includegraphics[width=\textwidth]{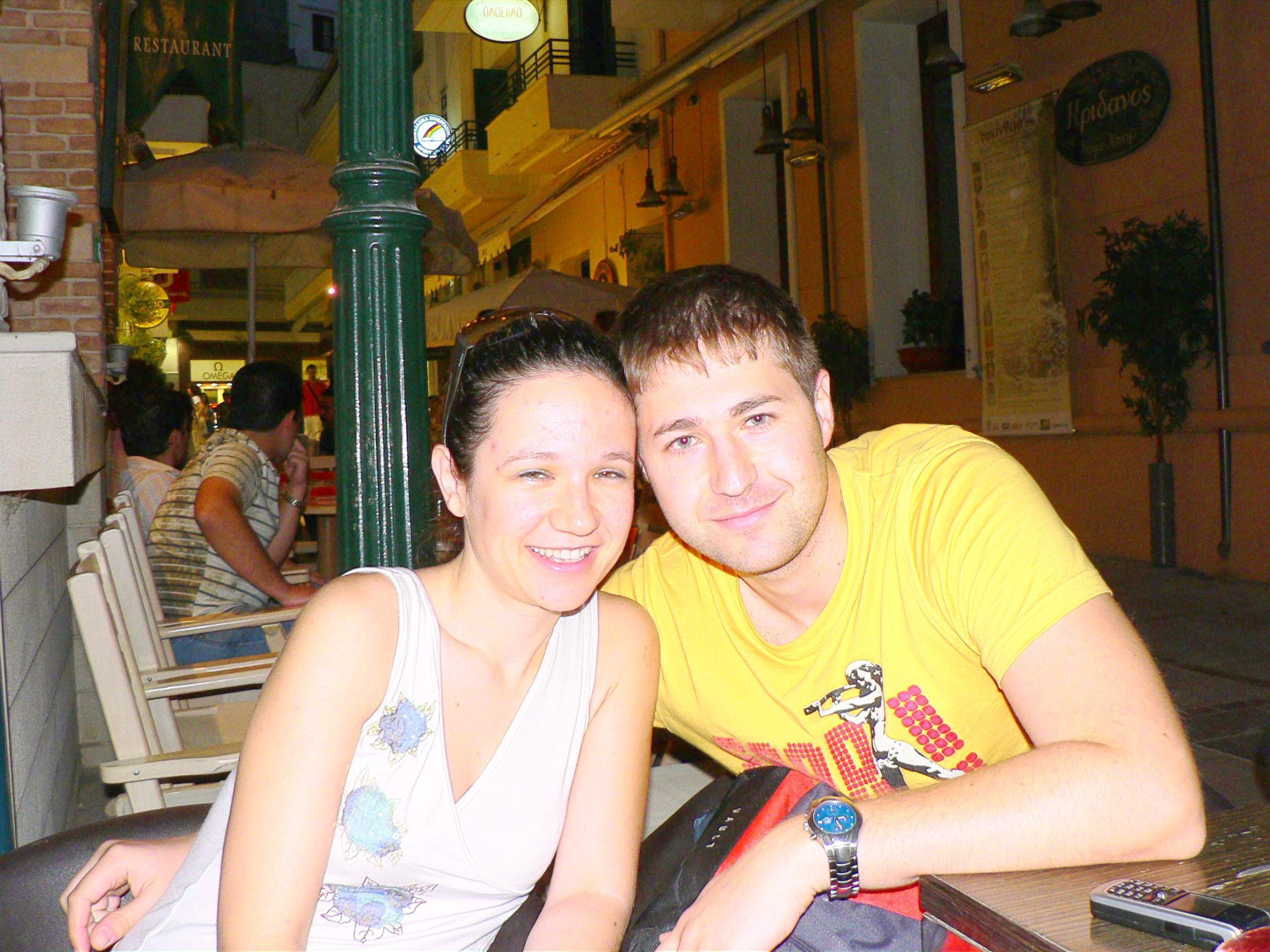}
			\caption{Inv-EnNet}
			\label{main_ablation_ours}
		\end{subfigure}
		\vspace{-3mm}
		\caption{Visual comparison of the results on VV to evaluate the various loss functions.}
		\label{fig:ablation_loss}
	\end{figure*}
	\section{Experiments}
	\label{sec:exp}
	\subsection{Datasets and Implementation Details}
	We use the training set in EnlightenGAN~\cite{jiang2021enlightengan}, which consists of 914 low-light images and 1016 normal-light photos. All images are resized to $600\times400$.
	For testing, we adopt the widely used benchmarks, including DICM~\cite{DICM}, LIME~\cite{LIME}, NPE~\cite{NPE}, MEF~\cite{MEF} and VV.
	The method is implemented with Pytorch on an NVIDIA 1080Ti GPU. Adam optimizer is used for training with the batch size of 16 and the learning rate of $1e-4$. Empirically, we set $\lambda=0.6$, $\mu=200$ and $\eta=0.5$.
	\subsection{Comparison with the state-of-the-arts}
	We perform evaluation with several current SOTAs on five benchmarks. Both the quantitative and qualitative results are provided for comparison.
	Table~\ref{tab:sota} reports the results of NIQE~\cite{niqe}, that is a non-reference image quality assessment metric widely used for unsupervised low-light enhancement~\cite{jiang2021enlightengan, kind++, llie_survey}.
	Notably, our proposed Inv-EnNet performs in top three on most of the datasets and achieves best result in average.
	In addition, an example of visual comparison is given in Figure~\ref{fig:sota}.
	In comparison with DeepUPE, our model reveals the details in the dark area of the low-light image significantly.
	Compared with EnlightenGAN and RUAS, the overexposed problem is effectively suppressed, which should be attributed to our designed reversibility loss considerably.
	Moreover, our approach can generate natural and vivid images without problems such as halo artifact in ZeroDCE.
	In Figure~\ref{fig:sota_unsupervised}, we further provide the visual comparison with the current state-of-the-art unsupervised methods on several challenging low-light images from NPE dataset.
	Note that, our result exhibits distinct contrast and more vivid color, while the compared methods suffer from severe artifacts.
	More visualization results can be found in the supplementary material.
	\begin{table}[htbp]
		\small
		\setlength\tabcolsep{.8pt}
		\renewcommand\arraystretch{1}
	\caption{Comparison in terms of NIQE with the SOTAs. Top three results are in \textcolor{red}{red}, \textcolor{blue}{blue} and \textcolor{green}{green} respectively.}
	\label{tab:sota}
	\vspace{-2mm}
	\begin{center}
		\begin{tabular}{|c|c|c|c|c|c|c|}
			\hline
			Methods &  \ \  DICM  \ \ &  \ \ MEF  \ \ &  \ \ LIME  \ \ &  \ \ NPE \ \  & \ \  VV  \ \ &  \ \ Avg. \ \   \\
			\hline
			Input& 4.2541& 4.2666& 4.3514& 4.3121& 3.5258& 4.1254\\
			RetinexNet& 4.4503& 4.3952& 4.5907& 4.5652& \textcolor{red}{2.6933}& 4.1187\\
			DeepUPE& 3.8907& 3.5263& 3.9393& \textcolor{green}{3.9838}& \textcolor{green}{2.9974}& 3.6760\\
			EnlightenGAN& \textcolor{blue}{3.5521}& \textcolor{red}{3.2307}& \textcolor{red}{3.6553}& 4.1129& 4.1234& \textcolor{green}{3.6640}\\
			KinD++& 3.7850& 3.7368& 4.7217& 4.3796& \textcolor{blue}{2.7016}& 3.6817\\
			ZeroDCE& \textcolor{green}{3.5642}& \textcolor{blue}{3.2829}& \textcolor{blue}{3.7686}& \textcolor{blue}{3.9263}& 3.2130& \textcolor{blue}{3.4969}\\
			DRBN stage1& 4.1854& 4.0852& 4.3007& 4.1532& 3.1762& 3.9819\\
			DRBN stage2& 4.2878& 4.2527& 4.4326& 4.2271& 3.5441& 4.1457\\
			RUAS& 4.7181& 3.8297& 4.2463& 5.5342& 4.6140& 4.5897\\
			\hline
			Inv-EnNet (Ours)& \textcolor{red}{3.5218}& \textcolor{green}{3.4189}& \textcolor{blue}{3.7686}& \textcolor{red}{3.8768}& 3.0042& \textcolor{red}{3.4498}\\
			\hline
		\end{tabular}
	\end{center}
	\vspace{-8mm}
\end{table}
\subsection{Ablation Study}
To investigate the proposed method, we conduct ablation study for the split strategy, the loss functions and the number of coupling layers.
All the results are listed in Table~\ref{tab:ablation}.

\noindent \textbf{Split Strategy.}
By adopting the squeeze layer, we obtain four sub-images and split them into two parts as input. Here, different split strategy is investigated.
The variant "$n$-split" in Table~\ref{tab:ablation} refers to assign $n$ sub-images to $\mathbf{x}^{0}_{a}$ and $\mathbf{y}^{K}_{a}$.
A balanced split strategy can improve the qualitative performance of our model.
As is shown in supplementary material, 1-split and 3-split
strategies are likely to produce results with artifacts and make
the enhanced images less visually pleasing. Thus, we choose
2-split strategy as the default setting of our model.
Considering the simplicity of operation and integration with testing phase, we take the way of average split.

\noindent \textbf{Loss Functions.}
To analyze the effect of different loss functions, we remove each component and provide the comparison results of corresponding variants in Table~\ref{tab:ablation} and Figure~\ref{fig:ablation_loss}.
Results show that the designed losses contribute to the performance improvement in terms of NIQE or visualization.
From Table~\ref{tab:ablation}, the NIQE drops when getting rid of $\mathcal{L}_{TC}$ or $\mathcal{L}_{DP}$.
Without $\mathcal{L}_{R}$, although the quantitative results are competitive, however the visual results suffer from over-exposure problem as shown in Figure~\ref{fig:ablation_loss}. 
Moreover, we also observe that $(i)$ removing $\mathcal{L}_{TC}$ reduces the qualitative performance as the stability of model drops, $(ii)$ without $\mathcal{L}_{DP}$, the enhanced result loses some details and suffers from color distortion, $(iii)$ $\mathcal{L}_{R}$ indeed suppresses the overexposure problem effectively.

\noindent \textbf{Number of Coupling Layers.}
First, we visualize the enhanced image by directly composing the output sub-images in forward process using squeeze layer shown in Figure~\ref{fig:framework} and find that the result contains obvious checkerboard artifact.
In Table~\ref{tab:ablation}, we evaluate the effect of the number of deployed coupling layers for our progressive enhancement.
It can be noted that increasing the number of coupling layers can improve the NIQE results.
For a trade-off with efficiency, we set $K=8$ as default in all experiments.
\begin{table}[htbp]
	\small
	\setlength\tabcolsep{.8pt}
	\renewcommand\arraystretch{1}
\caption{Results of ablation study in terms of NIQE.}
\label{tab:ablation}
%\vspace{-2mm}
\begin{center}
	\begin{tabular}{|c|c|c|c|c|c|c|}
		\hline
		\ \ \ \	Datasets \ \ \ \ & \ \ DICM \ \ & \ \ MEF \ \ & \ \ LIME \ \ & \ \ NPE \ \ & \ \ VV \ \ & \ \ Avg. \ \  \\
		\hline
		1-split & 3.6265& 3.6118& 3.8171& 3.7672& 3.0657& 3.5397\\
		3-split & 3.6005& 3.4630& 3.5660& 3.7470& 3.0895& 3.4885\\
		\hline
		$\text{Inv-EnNet}^{w/o}_{\mathcal{L}_{TC}}$ & 3.8118& 3.3868& 3.6005& 3.5579& 3.5480& 3.6679\\
		$\text{Inv-EnNet}^{w/o}_{\mathcal{L}_{DP}}$ & 3.8939& 3.6180& 3.8209& 4.2126& 3.5719& 3.8078\\
		$\text{Inv-EnNet}^{w/o}_{\mathcal{L}_{R}}$ & 3.5104& 3.3029& 3.5321& 3.7363& 3.0967& {3.4175}\\
		\hline
		K=4 & 3.6123& 3.4512& 3.8227& 3.8434& 3.0629& 3.5150\\
		K=12 & 3.5403& 3.4593& 3.8061& 3.7106& 2.9219& {3.4411}\\
		\hline
		Inv-EnNet & 3.5218& 3.4189& 3.7686& 3.8768& 3.0042& {3.4498}\\
		\hline
	\end{tabular}
\end{center}
\vspace{-8mm}
\end{table}

\begin{figure}[htbp]
\centering
\begin{subfigure}[b]{0.15\textwidth}
	\includegraphics[width=\textwidth]{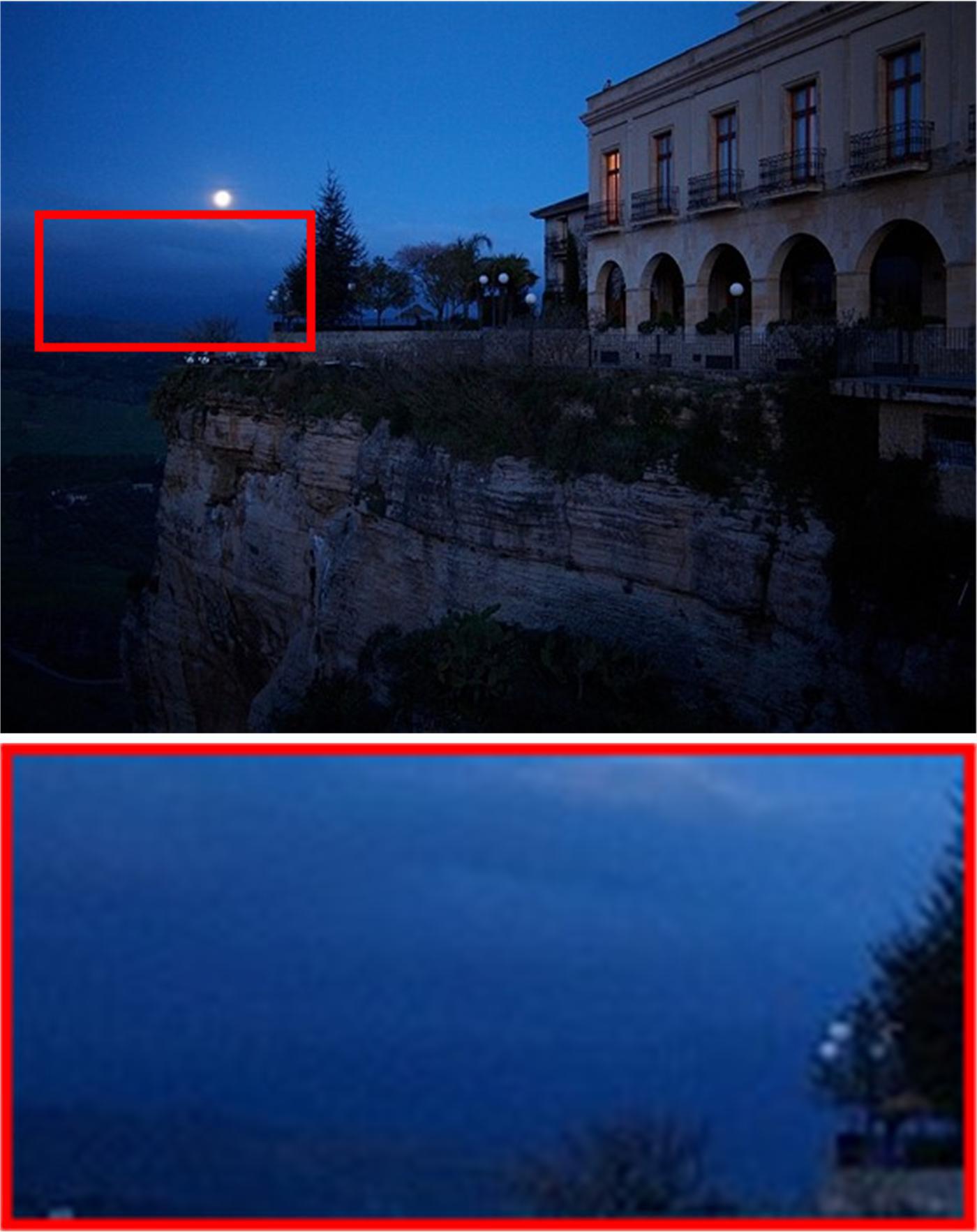}
	\caption{Input}
\end{subfigure}
\begin{subfigure}[b]{0.15\textwidth}
	\includegraphics[width=\textwidth]{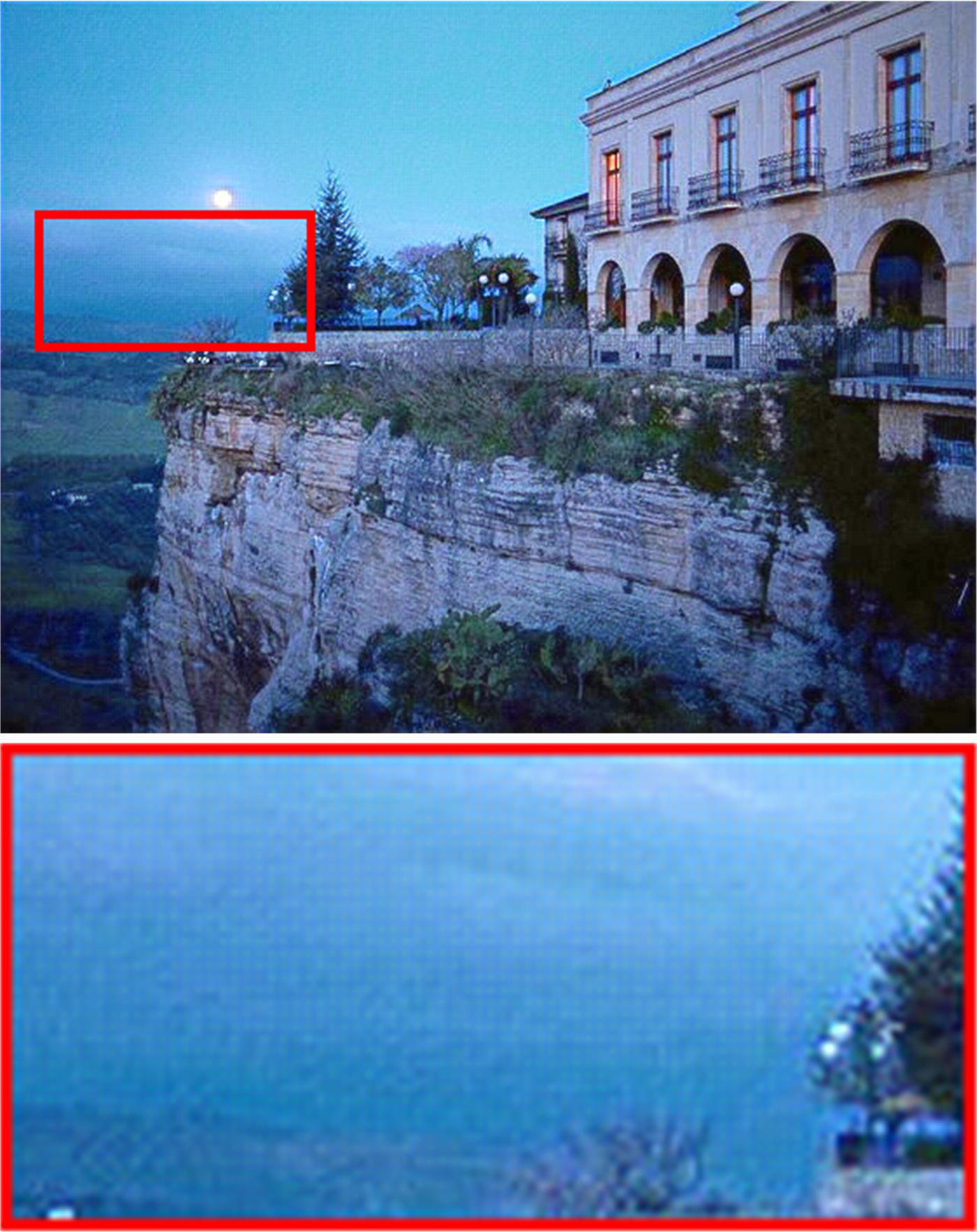}
	\caption{Unsqueeze}
\end{subfigure}
\begin{subfigure}[b]{0.15\textwidth}
	\centering
	\includegraphics[width=\textwidth]{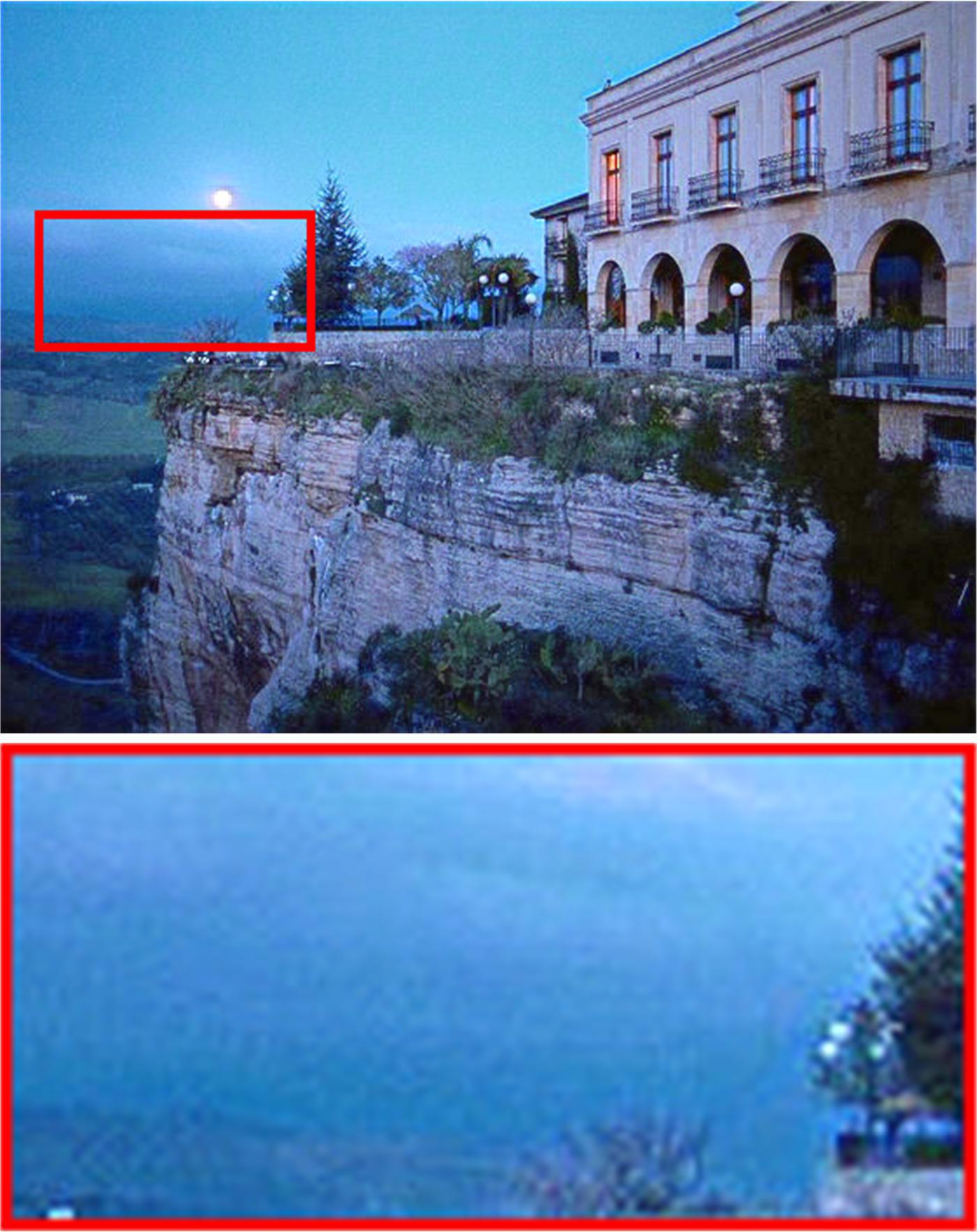}
	\caption{Inv-EnNet (Ours)}
\end{subfigure}
\vspace{-4mm}
\caption{Directly unsqueezing the sub-images $\{\hat{\alpha}_i\}_{i=0}^3$ leads to the checkerboard artifact as shown in (b). However, our progressive self-guided enhancement can mitigate this problem and produce visually pleasing result in (c).}
\label{fig:unsqueeze}
\vspace{-6mm}
\end{figure}
\section{Conclusion}
\label{sec:conc}
In this paper, we propose an invertible network based method for unpaired learning of the low-light image enhancement.
Inv-EnNet makes the training more stable than those two-way GAN frameworks.
To prevent from the checkerboard artifacts caused by squeeze layers, we present a progressive self-guided enhancement process to enhance the low-light images.
Moreover, to produce visually pleasing results, we also design various loss functions, especially the revertible loss to mitigate the overexposure problem.
Experimental results demonstrate the superiority of our method against existing low-light enhancement methods both in terms of quantification and visualization.

\bibliographystyle{IEEEbib}
\bibliography{Inv-EnNet}

\end{document}

\sloppy

\def\x{{\mathbf x}}
\def\L{{\cal L}}

\title{Invertible Network for Unpaired Low-Light Image Enhancement}

	\name{Jize Zhang, Haolin Wang, Xiaohe Wu and Wangmeng Zuo}
	\address{School of Computer Science and Technology, Harbin Institute of Technology, China \\ 
	 jize.zhang.cs@outlook.com,
	 Why\_cs@outlook.com,
	 xhwu.cpsl.hit@gmail.com,
	 cswmzuo@gmail.com}

\maketitle

\section{Detailed Network Structure}

\begin{figure}[htbp]
    \centering
    \includegraphics[width=0.9\columnwidth]{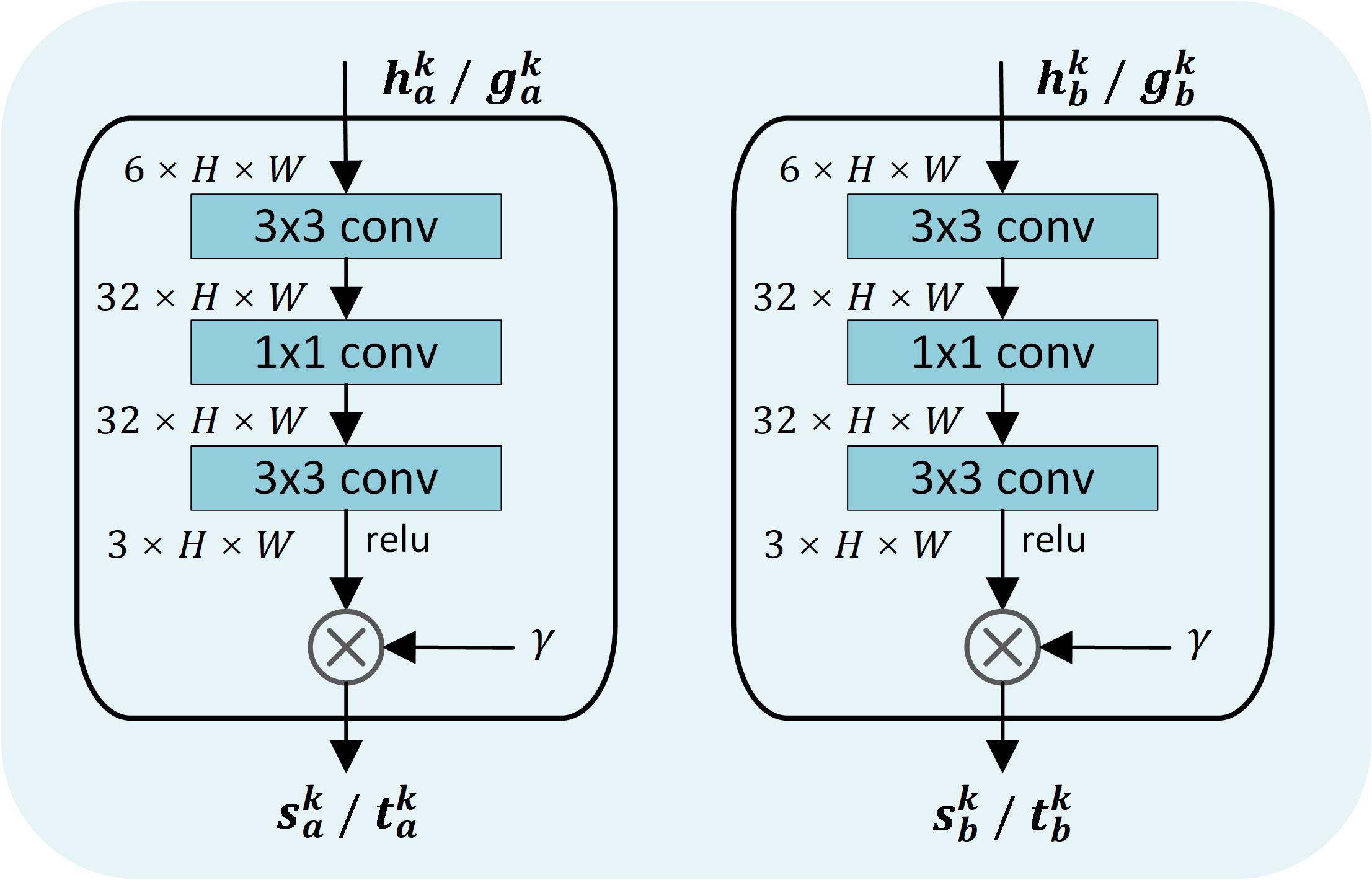}
    \caption{The detailed structure of our non-linear mappings in each affine coupling layers.}
    
    \label{subnet}
    \vspace{-8mm}
\end{figure}

In Figure~\ref{subnet}, we present the detailed structure of non-linear mappings for the translations, i.e., $h^k_a(\cdot)$, $h^k_b(\cdot)$, $g^k_a(\cdot)$ and $g^k_b(\cdot)$ respectively. 
As shown, each mapping consists of 3 convolutional layers followed by a relu activation function. 
Then, we multiply the output with the trainable scalar parameter $\gamma$.
The shape of all the output $s_a^k$, $s_b^k$, $t_a^k$ and $t_b^k$ is $3\times H\times W$.

Following EnlightenGAN~\cite{jiang2021enlightengan}, we adopt the PatchGAN as our discriminators for low-light image discriminator $D^L$ and the normal-light image discriminator $D^N$. 
Detailed structure of the discriminators is shown in Table~\ref{tab:discriminator}. 

\section{Ablation Study Results}

In this section, we provide more results of ablation study, including the impact of split strategy and the number of coupling layers. We also illustrate the superiority of our INN-based framework over its forward-only variant.

\begin{figure*}
    \centering
    \begin{subfigure}[b]{0.24\textwidth}
        \centering
        \includegraphics[width=\textwidth]{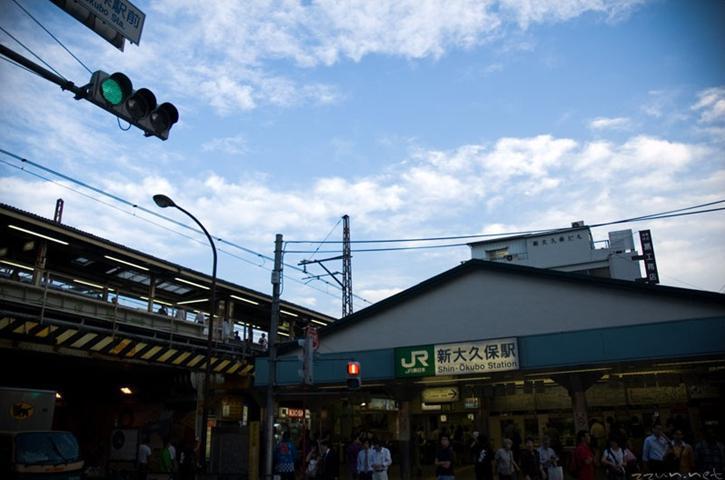}
        \caption{Input}
        \label{split_in}
    \end{subfigure}
    \hfill
    \begin{subfigure}[b]{0.24\textwidth}
        \centering
        \includegraphics[width=\textwidth]{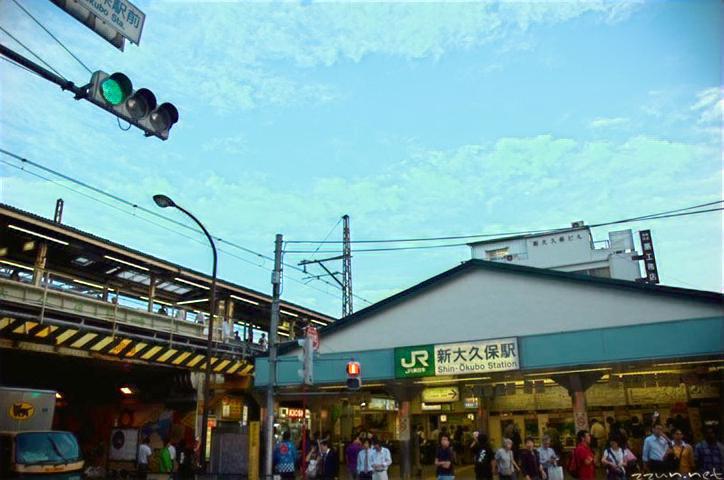}
        \caption{1-split}
        \label{1split}
    \end{subfigure}
    \hfill
    \begin{subfigure}[b]{0.24\textwidth}
    \centering
        \includegraphics[width=\textwidth]{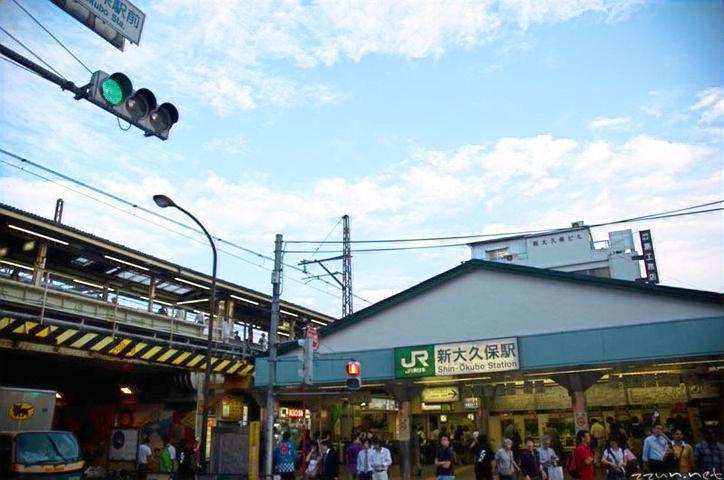}
        \caption{{2-split} (Default) }
        \label{2split}
    \end{subfigure}
    \begin{subfigure}[b]{0.24\textwidth}
    \centering
        \includegraphics[width=\textwidth]{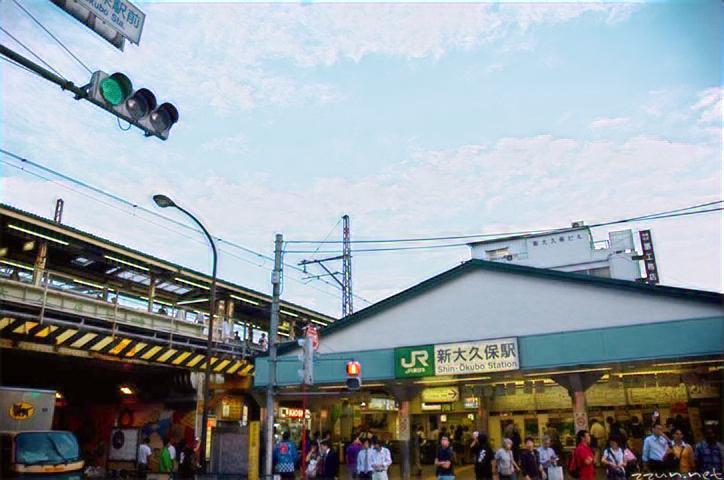}
        \caption{3-split}
        \label{3split}
    \end{subfigure}
        \caption{Ablation study of the split strategy. The 2-split scheme is the default setting.}
        \label{fig:split}
\end{figure*}

\noindent \textbf{Split Strategy.} As shown in Figure~\ref{fig:split}, 1-split and 3-split strategies are likely to produce results with artifacts and make the enhanced images less visually pleasing. 
Thus, we choose 2-split strategy as the default setting of our model.

\begin{figure*}
    \centering
    \begin{subfigure}[b]{0.19\textwidth}
        \centering
        \includegraphics[width=\textwidth]{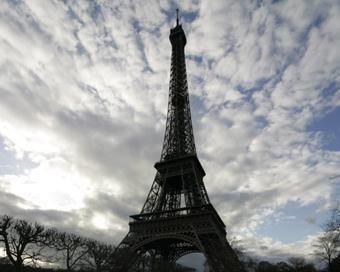}
        \caption{Input}
    \end{subfigure}
    \hfill
    \begin{subfigure}[b]{0.19\textwidth}
    \centering
        \includegraphics[width=\textwidth]{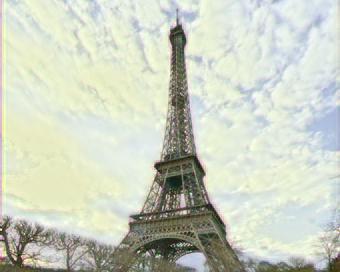}
        \caption{$\text{Inv-EnNet}^{w/o}_{\mathcal{L}_{TC}}$}
    \end{subfigure}
    \hfill
    \begin{subfigure}[b]{0.19\textwidth}
        \centering
        \includegraphics[width=\textwidth]{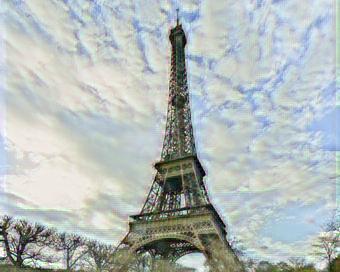}
        \caption{$\text{Inv-EnNet}^{w/o}_{\mathcal{L}_{DP}}$}
    \end{subfigure}
    \hfill
    \begin{subfigure}[b]{0.19\textwidth}
    \centering
        \includegraphics[width=\textwidth]{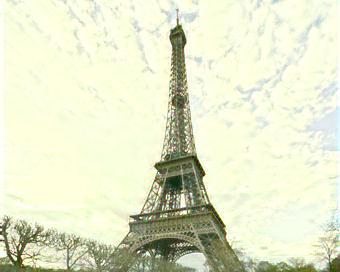}
        \caption{$\text{Inv-EnNet}^{w/o}_{\mathcal{L}_{R}}$}
    \end{subfigure}
    \begin{subfigure}[b]{0.19\textwidth}
    \centering
        \includegraphics[width=\textwidth]{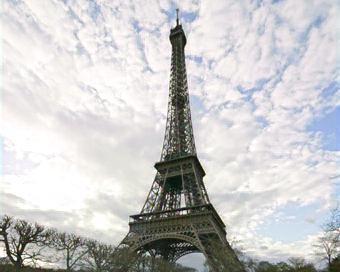}
        \caption{$\text{Inv-EnNet}$}
    \end{subfigure}
        \caption{Ablation study of the loss functions.}
        \label{fig:ex_loss}
\end{figure*}

\noindent \textbf{Loss Functions.}
To analyze the effect of different loss functions, we remove each component and provide more comparison results. As shown in Figure~\ref{fig:ex_loss}, $(i)$ removing $\mathcal{L}_{TC}$ reduces the qualitative performance, $(ii)$ without $\mathcal{L}_{DP}$, the enhanced results cannot preserve details and also exist color distortion, $(iii)$ $\mathcal{L}_{R}$ alleviates the over-exposure issue effectively.

\begin{figure*}
    \centering
    \begin{subfigure}[b]{0.24\textwidth}
        \centering
        \includegraphics[width=\textwidth]{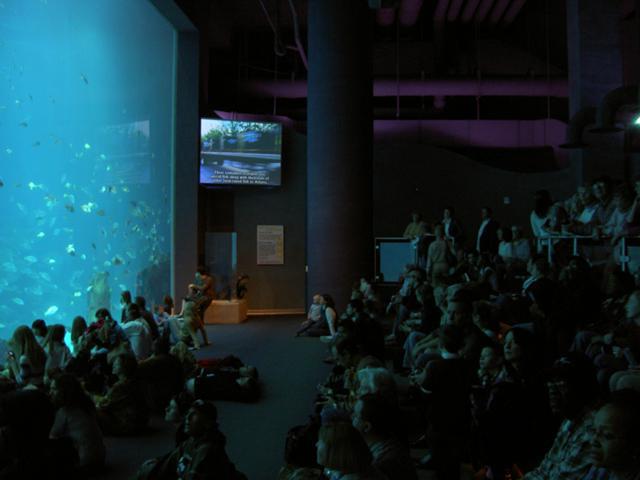}
        \caption{Input}
        \label{layer_num_in}
    \end{subfigure}
    \hfill
    \begin{subfigure}[b]{0.24\textwidth}
        \centering
        \includegraphics[width=\textwidth]{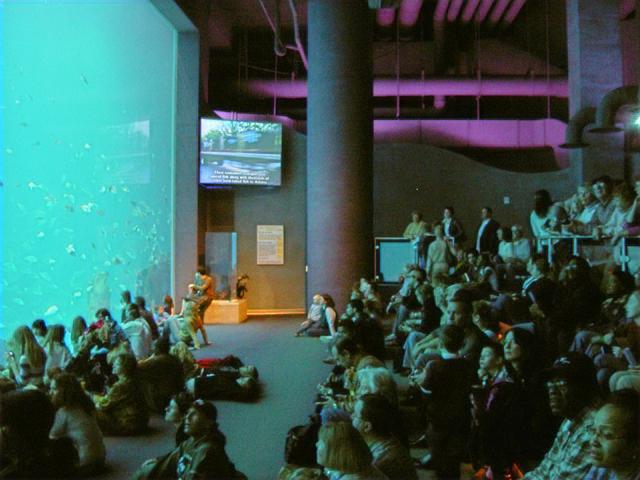}
        \caption{$K=4$}
        \label{K=4}
    \end{subfigure}
    \hfill
    \begin{subfigure}[b]{0.24\textwidth}
    \centering
        \includegraphics[width=\textwidth]{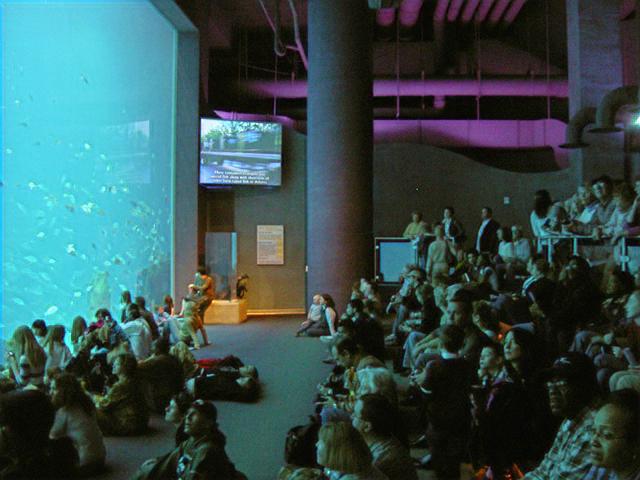}
        \caption{{$K=8$} (Default) }
        \label{K=8}
    \end{subfigure}
    \begin{subfigure}[b]{0.24\textwidth}
    \centering
        \includegraphics[width=\textwidth]{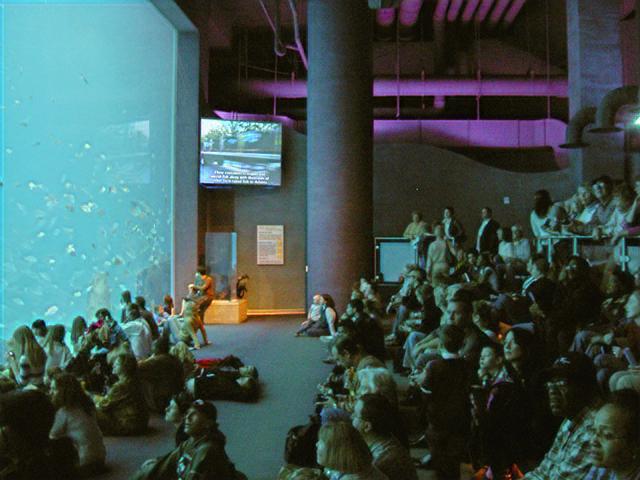}
        \caption{$K=12$}
        \label{K=12}
    \end{subfigure}
        \caption{Ablation study of the number of coupling layers. We use $K=8$ as default.}
        \label{fig:ablation_num}
\end{figure*}

\noindent \textbf{Number of Coupling Layers.} As shown in Figure~\ref{fig:ablation_num}, when setting the number of coupling layers $K$ as 4, the color of aquarium is less natural. In the meantime, the quantitative performance also drops from 3.4498 to 3.5150 in terms of NIQE. 
On the other hand, increasing $K$ from 8 to 12 produces a slightly improvement in qualitative and quantitative performance. Considering the memory and efficiency, we set $K=8$ as a trade-off in all experiments.

\begin{figure*}
    \begin{subfigure}[b]{0.32\textwidth}
        \centering
        \includegraphics[width=\textwidth]{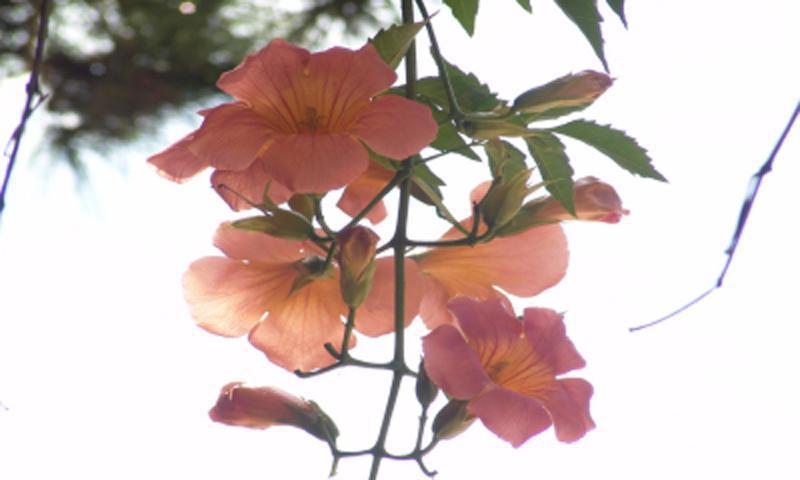}
        \end{subfigure}
    \hfill
    \begin{subfigure}[b]{0.32\textwidth}
        \centering
        \includegraphics[width=\textwidth]{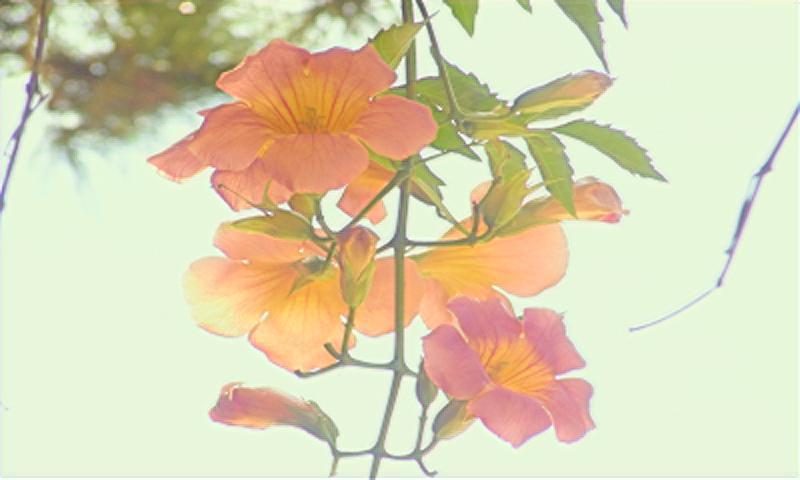}
        \end{subfigure}
    \hfill
    \begin{subfigure}[b]{0.32\textwidth}
    \centering
        \includegraphics[width=\textwidth]{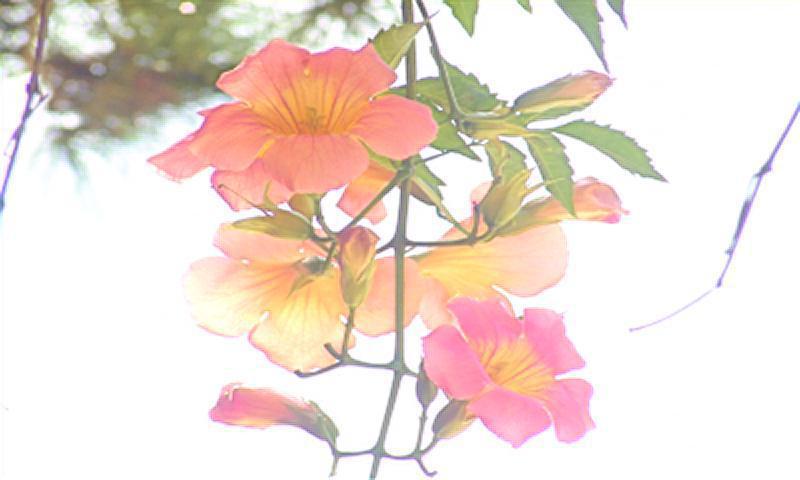}
        \end{subfigure}
    \centering
    \begin{subfigure}[b]{0.32\textwidth}
        \centering
        \includegraphics[width=\textwidth]{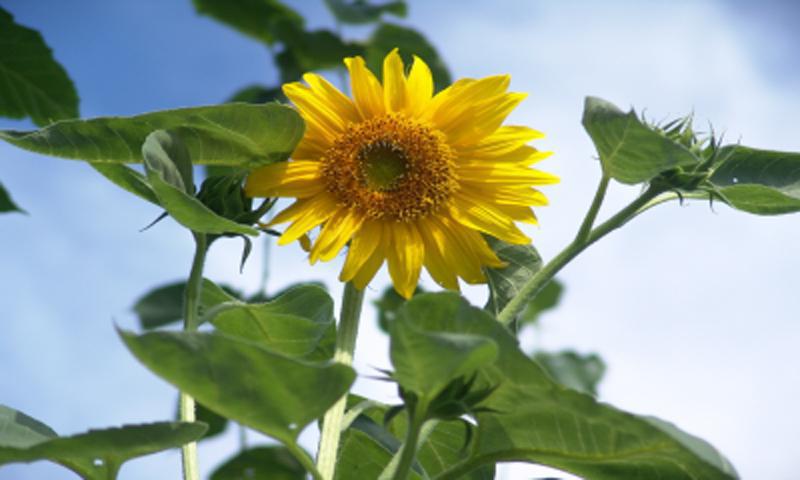}
        \caption{Input}
        \end{subfigure}
    \hfill
    \begin{subfigure}[b]{0.32\textwidth}
        \centering
        \includegraphics[width=\textwidth]{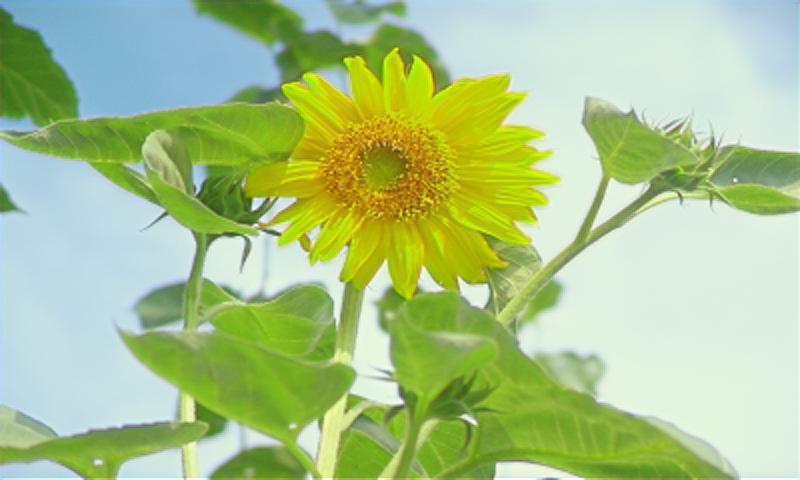}
        \caption{Forward-Only Inv-EnNet}
        \end{subfigure}
    \hfill
    \begin{subfigure}[b]{0.32\textwidth}
    \centering
        \includegraphics[width=\textwidth]{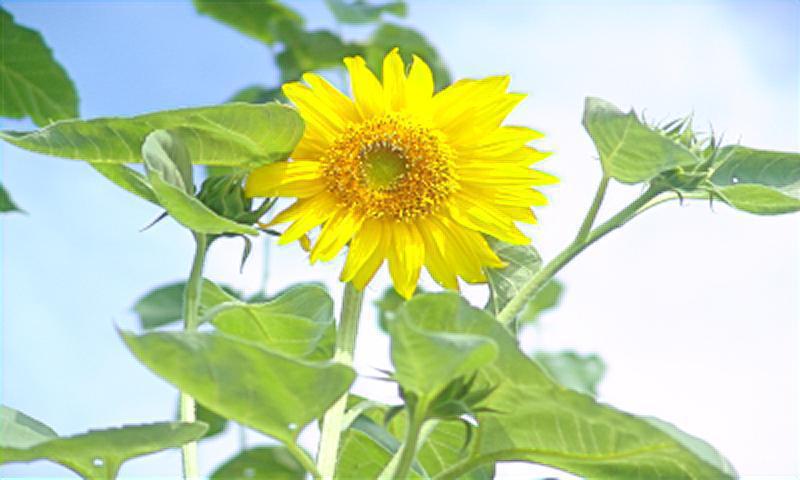}
        \caption{{Inv-EnNet}}
    \end{subfigure}
        \caption{Ablation study to compare with the Forward-Only variant of Inv-EnNet.}
        \label{fig:ablation_oneway}
\end{figure*}

\noindent \textbf{Forward-Only Inv-EnNet.} We also provide a variant of Inv-EnNet with only forward process, namely Forward-Only Inv-EnNet, to demonstrate the advantage of the invertible network. It implies that the unpaired normal-light images are used directly to train a discriminator for adversarial learning in combination with the enhanced results. From Table~\ref{tab:oneway} and Figure~\ref{fig:ablation_oneway}, results from Inv-EnNet could preserve the color from original image, and achieve better NIQE result compared with its Forward-Only variant.

\begin{table}[htbp]
\small
\caption{Structure of the discriminators $D^N$ and $D^L$.}
\vspace{1mm}
\label{tab:discriminator}
\begin{center}
	\begin{tabular}{|c|c|c|c|c|}
		\hline
		Layer& Kernel& stride& $C_{in}$& $C_{out}$\\
		\hline
		Conv + LReLU& 4$\times$4 & 2 & 2 & 64\\
        Conv + LReLU& 4$\times$4 & 2 & 64 & 128\\
        Conv + LReLU& 4$\times$4 & 2 & 128 & 256\\
        Conv + LReLU& 4$\times$4 & 2 & 256 & 512\\
        Conv + LReLU& 4$\times$4 & 2 & 512 & 512\\
        Conv + LReLU& 4$\times$4 & 1 & 512 & 512\\
        Conv& 4$\times$4 & 1 & 512 & 1\\
		\hline
	\end{tabular}
\end{center}
\vspace{-12mm}
\end{table}

\iffalse
\begin{table}[htbp]
	\small
	\setlength\tabcolsep{.8pt}
	\renewcommand\arraystretch{1}
\caption{Results of ablation study in terms of NIQE.}
\label{tab:oneway}
\begin{center}
	\begin{tabular}{|c|c|c|c|c|c|c|}
		\hline
    	\ \ \ \	Datasets \ \ \ \ & \ \ DICM \ \ & \ \ MEF \ \ & \ \ LIME \ \ & \ \ NPE \ \ & \ \ VV \ \ & \ \ Avg. \ \  \\
    	\hline
    	Forward-Only Inv-EnNet GAN & 3.8567 & 3.6089 & 3.6783 & 3.7498 & 3.3951 & 3.7109 \\
	    Inv-EnNet & 3.5218& 3.4189& 3.7686& 3.8768& 3.0042& {3.4498} \\
	    \hline
	\end{tabular}
\end{center}
\vspace{-8mm}
\end{table}
\fi

\begin{table}[htbp]
	\small
	\setlength\tabcolsep{.8pt}
	\renewcommand\arraystretch{1}
\caption{Results of ablation study in terms of NIQE.}
\label{tab:oneway}
\begin{center}
	\begin{tabular}{|c|c|c|}
		\hline
		\ \ \ \ Datasets \ \ \ \ & \ Forward-Only Inv-EnNet \  & \ \ Inv-EnNet \ \  \\
		\hline
		DICM &3.8567 &3.5218 \\
		MEF  &3.6089 &3.4189 \\
		LIME &3.6783 &3.7686 \\
		NPE  &3.7498 &3.8768 \\
		VV   &3.3951 &3.0042 \\
		\hline
		Avg. &3.7109 &{3.4498} \\
	    \hline
	\end{tabular}
\end{center}
\vspace{-10mm}
\end{table}

\section{Visual Comparison Results}
\begin{figure*}[htbp]
    \centering
    \begin{subfigure}[b]{0.19\textwidth}
        \centering
        \hspace{-1.6mm}
        \includegraphics[width=\textwidth]{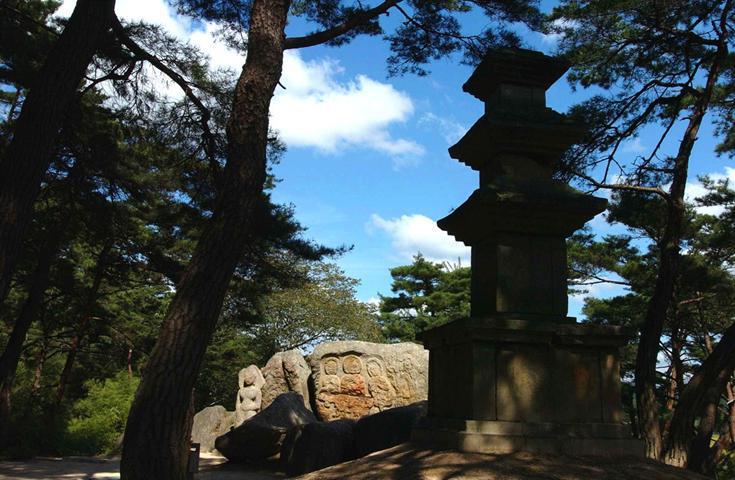}
        \caption{Input}
    \end{subfigure}
    \begin{subfigure}[b]{0.19\textwidth}
    \centering
    \hspace{-1.6mm}
    \includegraphics[width=\textwidth]{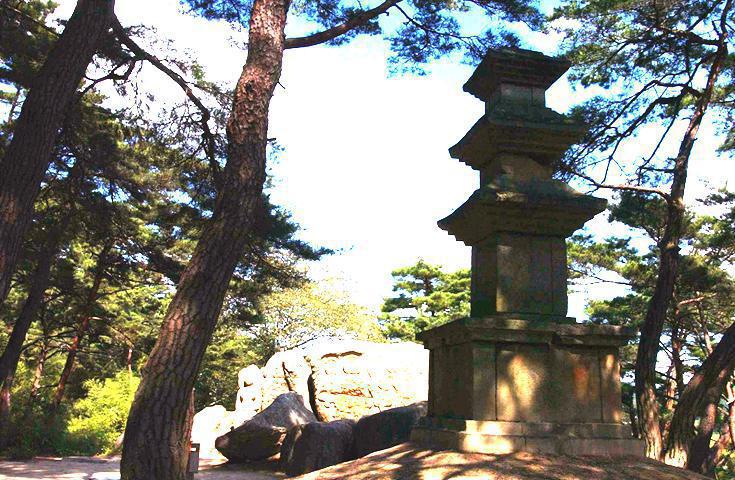}
        \caption{RUAS~\cite{liu2021ruas}}
    \end{subfigure}
    \begin{subfigure}[b]{0.19\textwidth}
        \centering
        \hspace{-1.6mm}
        \includegraphics[width=\textwidth]{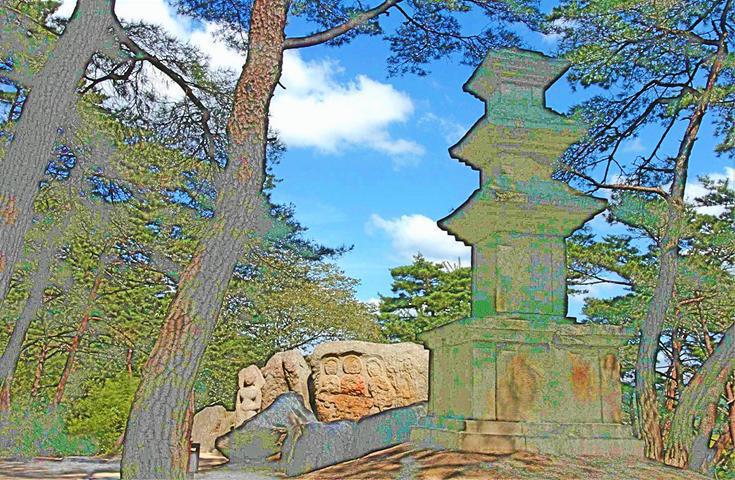}
        \caption{RetinexNet~\cite{Chen2018Retinex}}
    \end{subfigure}
    \begin{subfigure}[b]{0.19\textwidth}
        \centering
        \hspace{-1.6mm}
        \includegraphics[width=\textwidth]{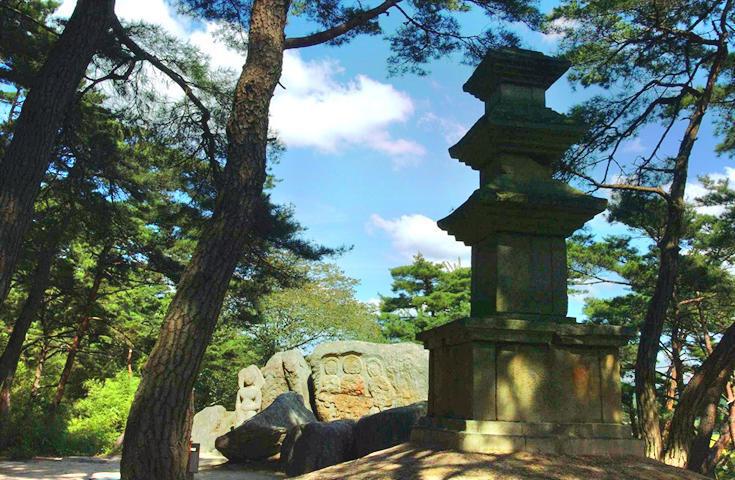}
        \caption{DeepUPE~\cite{Wang_2019_CVPR}}
    \end{subfigure}
    \begin{subfigure}[b]{0.19\textwidth}
    \centering
    \hspace{-1.6mm}
    \includegraphics[width=\textwidth]{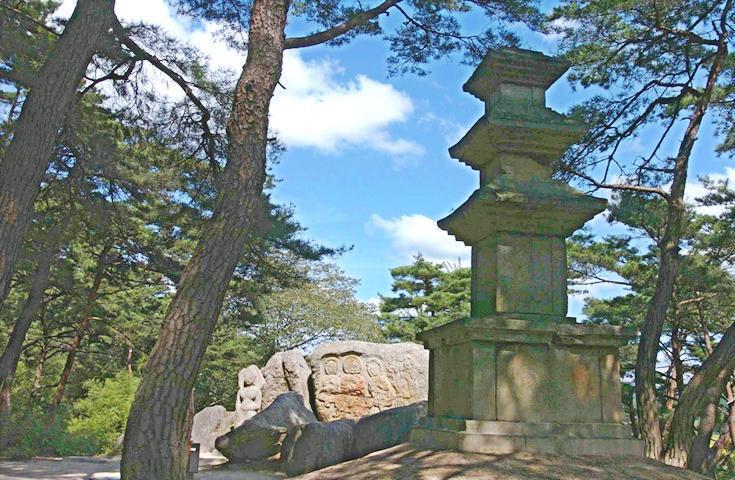}
        \caption{ZeroDCE~\cite{Zero-DCE}}
    \end{subfigure}
    \begin{subfigure}[b]{0.19\textwidth}
    \centering
        \includegraphics[width=\textwidth]{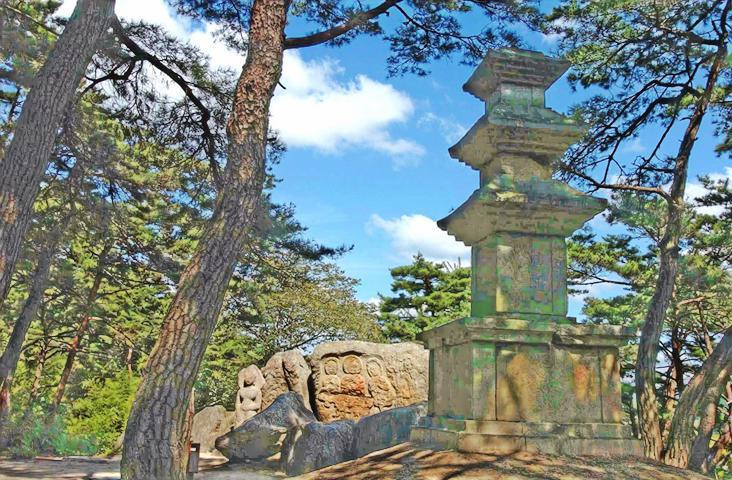}
        \caption{KinD++~\cite{kind++}}
    \end{subfigure}
    \begin{subfigure}[b]{0.19\textwidth}
        \centering
        \includegraphics[width=\textwidth]{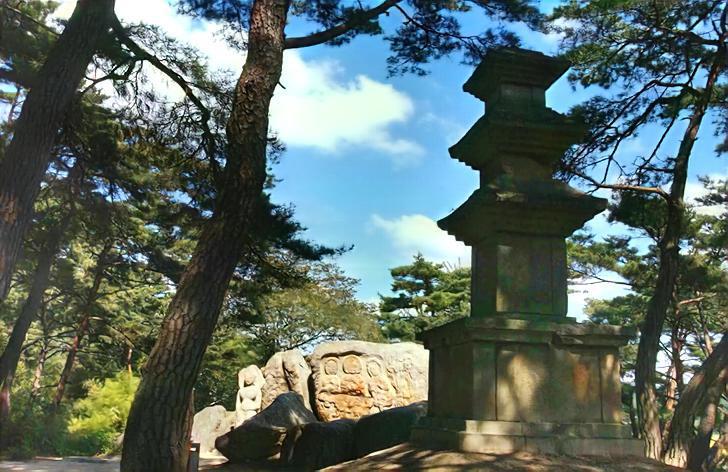}
        \caption{DRBN stage1~\cite{drbn}}
    \end{subfigure}
    \begin{subfigure}[b]{0.19\textwidth}
        \centering
        \includegraphics[width=\textwidth]{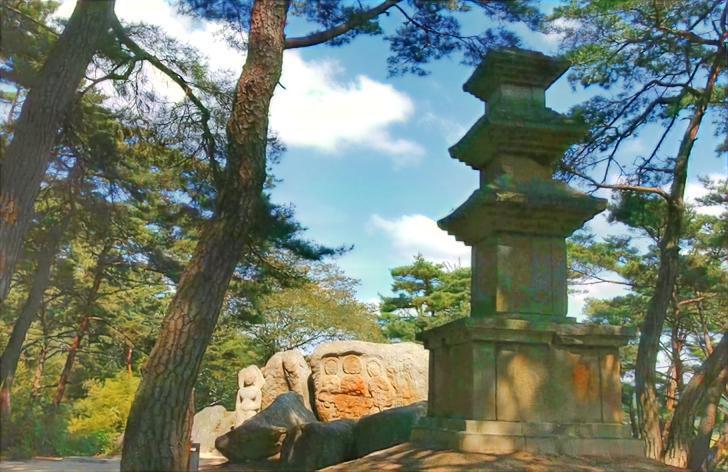}
        \caption{DRBN stage2~\cite{drbn}}
    \end{subfigure}
    \begin{subfigure}[b]{0.19\textwidth}
        \centering
        \includegraphics[width=\textwidth]{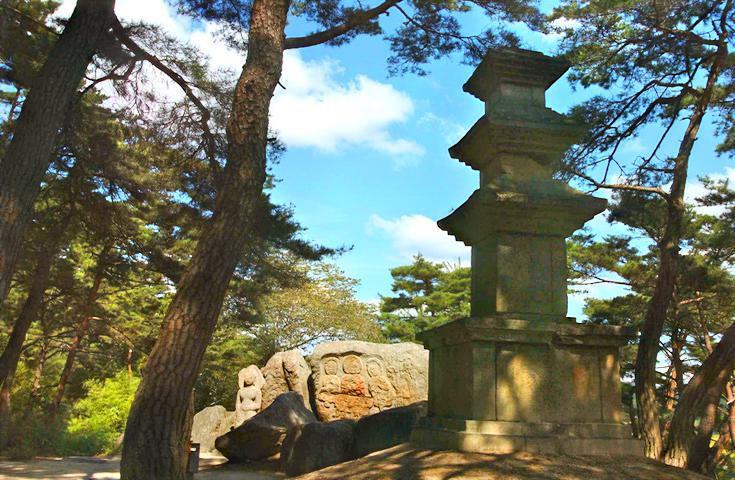}
        \caption{EnlightenGAN~\cite{jiang2021enlightengan}}
    \end{subfigure}
    \begin{subfigure}[b]{0.19\textwidth}
    \centering
        \includegraphics[width=\textwidth]{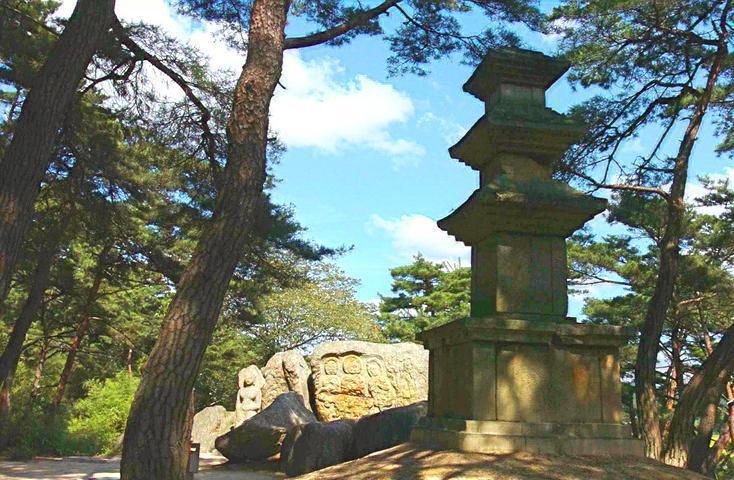}
        \caption{Inv-EnNet (Ours)}
    \end{subfigure}
    \vspace{-2mm}
    \caption{Visual comparison with SOTA methods on the DICM dataset.}
    \vspace{-2mm}
    \label{fig:sota1}
\end{figure*}

\begin{figure*}[htbp]
    \centering
    \begin{subfigure}[b]{0.19\textwidth}
        \centering
        \hspace{-1.6mm}
        \includegraphics[width=\textwidth]{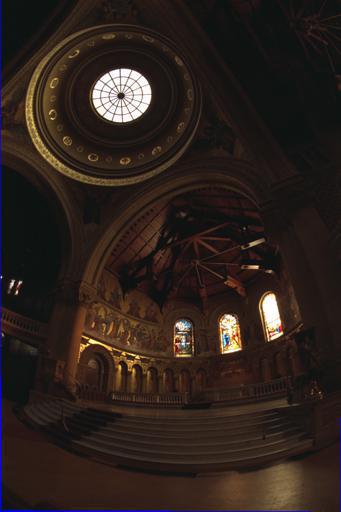}
        \caption{Input}
    \end{subfigure}
    \begin{subfigure}[b]{0.19\textwidth}
    \centering
    \hspace{-1.6mm}
    \includegraphics[width=\textwidth]{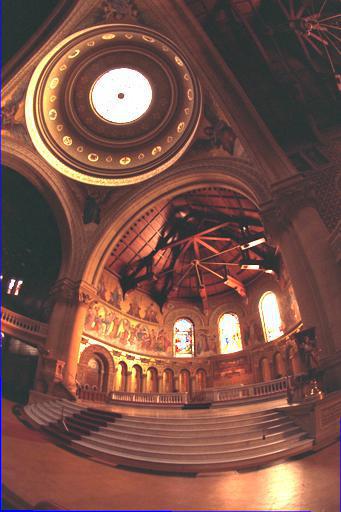}
        \caption{RUAS~\cite{liu2021ruas}}
    \end{subfigure}
    \begin{subfigure}[b]{0.19\textwidth}
        \centering
        \hspace{-1.6mm}
        \includegraphics[width=\textwidth]{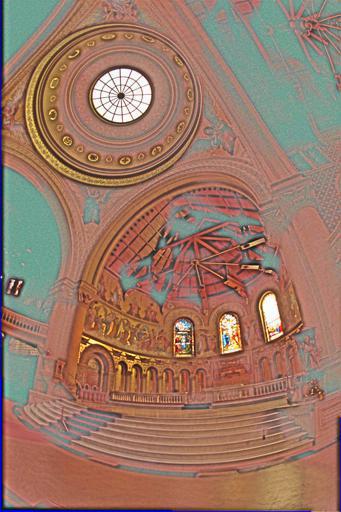}
        \caption{RetinexNet~\cite{Chen2018Retinex}}
    \end{subfigure}
    \begin{subfigure}[b]{0.19\textwidth}
        \centering
        \hspace{-1.6mm}
        \includegraphics[width=\textwidth]{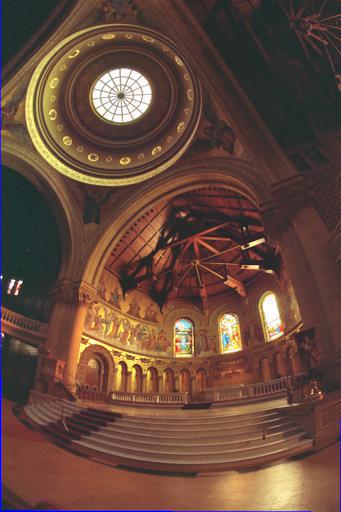}
        \caption{DeepUPE~\cite{Wang_2019_CVPR}}
        \label{qualitative_DeepUPE}
    \end{subfigure}
    \begin{subfigure}[b]{0.19\textwidth}
    \centering
    \hspace{-1.6mm}
    \includegraphics[width=\textwidth]{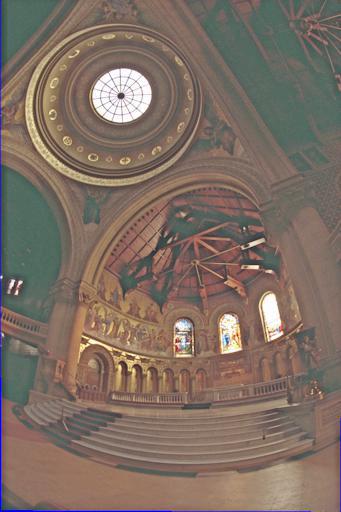}
        \caption{ZeroDCE~\cite{Zero-DCE}}
    \end{subfigure}
    \begin{subfigure}[b]{0.19\textwidth}
    \centering
        \includegraphics[width=\textwidth]{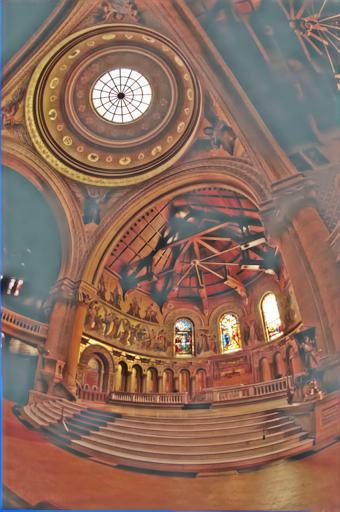}
        \caption{KinD++~\cite{kind++}}
    \end{subfigure}
    \begin{subfigure}[b]{0.19\textwidth}
        \centering
        \includegraphics[width=\textwidth]{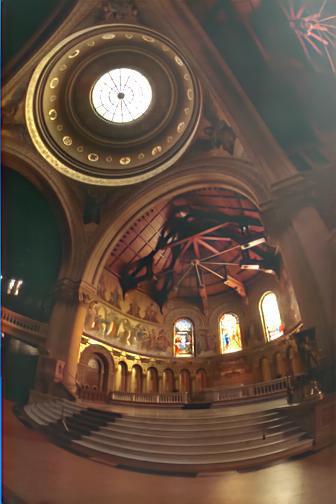}
        \caption{DRBN stage1~\cite{drbn}}
    \end{subfigure}
    \begin{subfigure}[b]{0.19\textwidth}
        \centering
        \includegraphics[width=\textwidth]{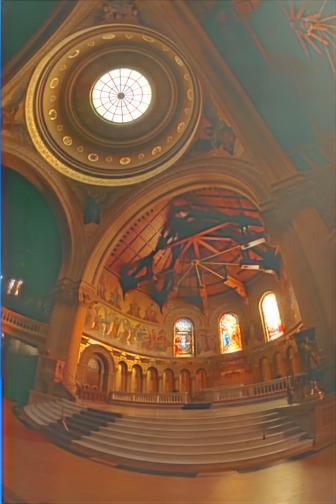}
        \caption{DRBN stage2~\cite{drbn}}
    \end{subfigure}
    \begin{subfigure}[b]{0.19\textwidth}
        \centering
        \includegraphics[width=\textwidth]{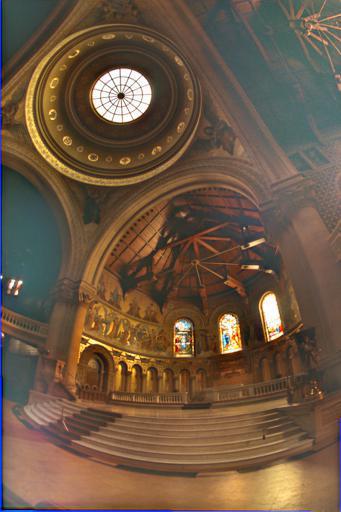}
        \caption{EnlightenGAN~\cite{jiang2021enlightengan}}
    \end{subfigure}
    \begin{subfigure}[b]{0.19\textwidth}
    \centering
        \includegraphics[width=\textwidth]{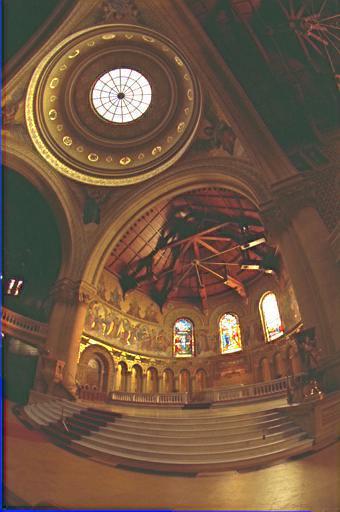}
        \caption{Inv-EnNet (Ours)}
    \end{subfigure}
    \vspace{-2mm}
    \caption{Visual comparison with SOTA methods on the MEF dataset.}
    \vspace{-16mm}
    \label{fig:sota3}
\end{figure*}

In this section, we show more visual comparison results in Figure~\ref{fig:sota1} and \ref{fig:sota3}. Our method generates results with high visual quality and vivid color, while the compared methods generate ones with severe artifacts.

\begin{figure*}[htbp]
    \centering
    \begin{subfigure}[b]{0.19\textwidth}
        \centering
        \hspace{-1.6mm}
        \includegraphics[width=\textwidth]{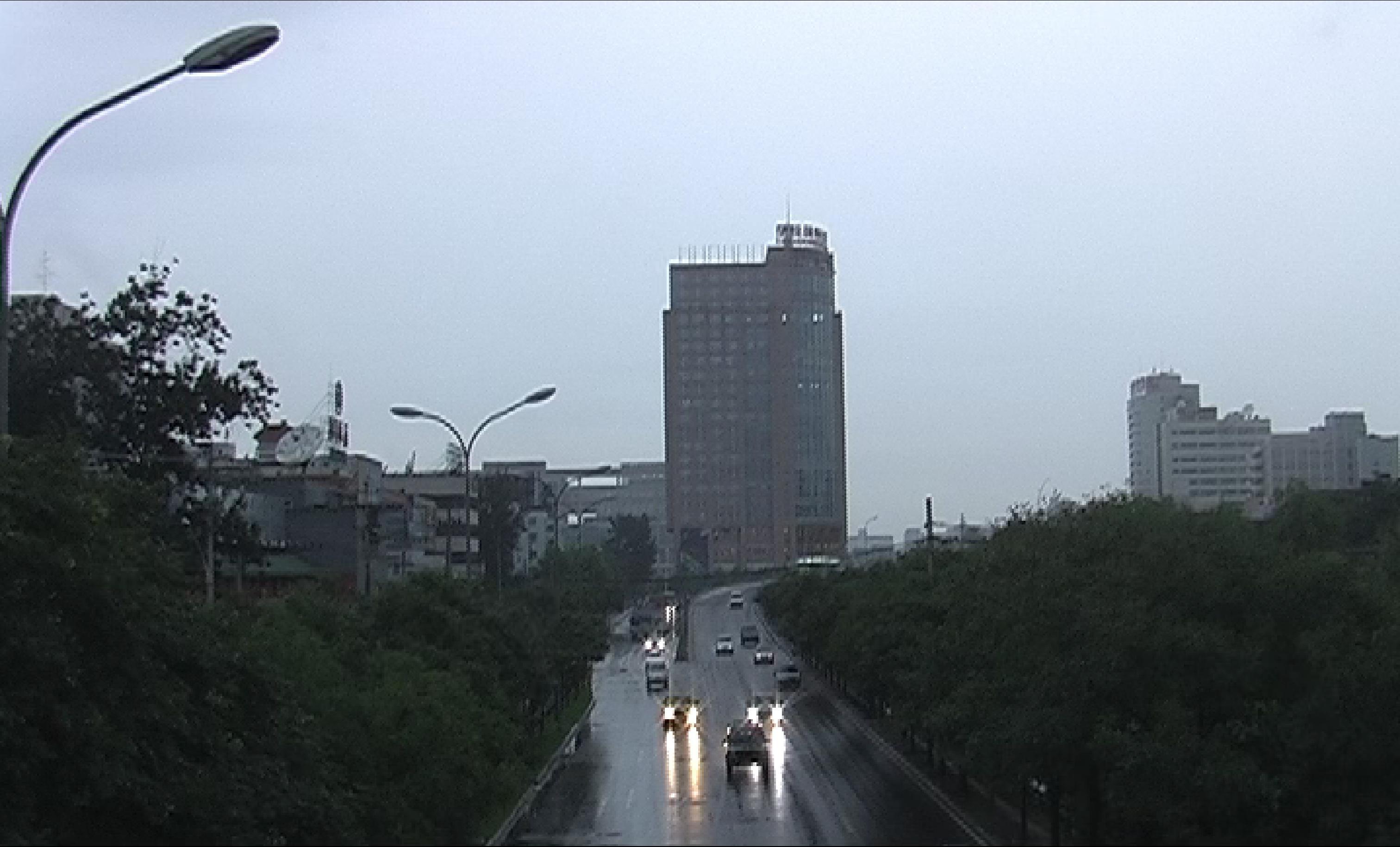}
        \caption{Input}
    \end{subfigure}
    \begin{subfigure}[b]{0.19\textwidth}
    \centering
    \hspace{-1.6mm}
    \includegraphics[width=\textwidth]{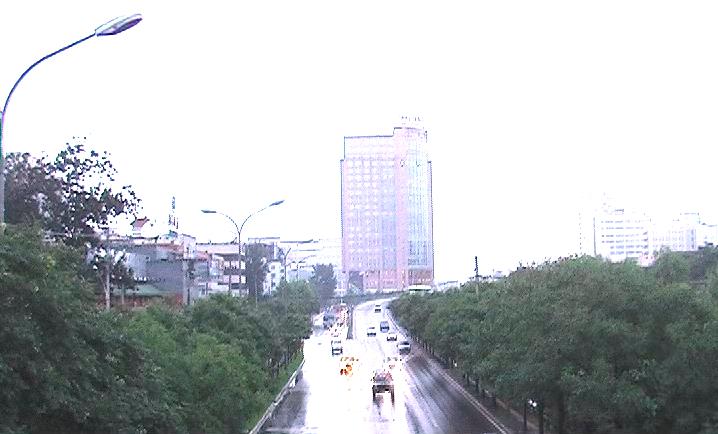}
        \caption{RUAS~\cite{liu2021ruas}}
    \end{subfigure}
    \begin{subfigure}[b]{0.19\textwidth}
        \centering
        \hspace{-1.6mm}
        \includegraphics[width=\textwidth]{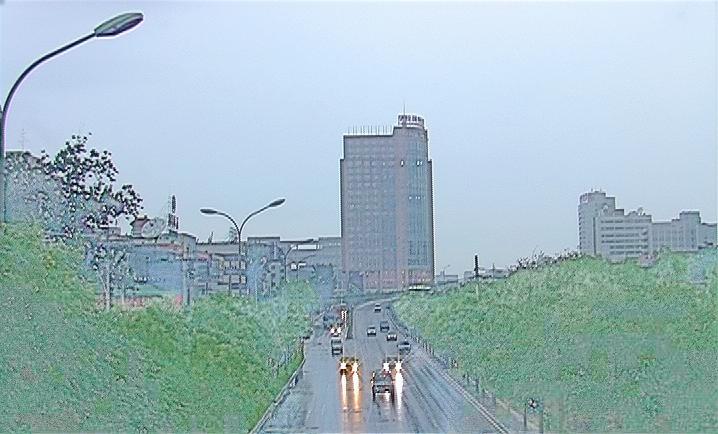}
        \caption{RetinexNet~\cite{Chen2018Retinex}}
    \end{subfigure}
    \begin{subfigure}[b]{0.19\textwidth}
        \centering
        \hspace{-1.6mm}
        \includegraphics[width=\textwidth]{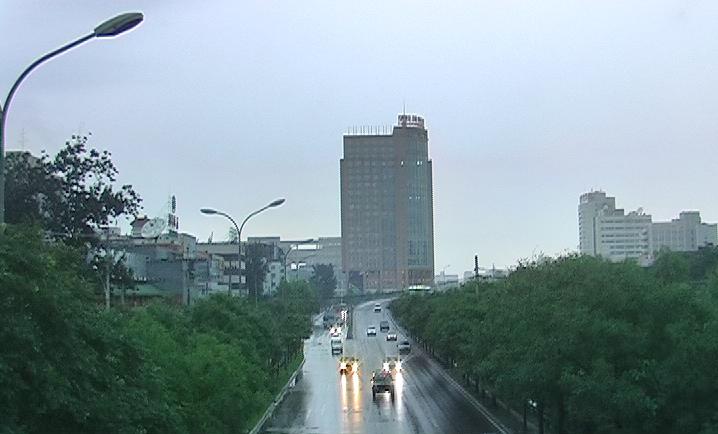}
        \caption{DeepUPE~\cite{Wang_2019_CVPR}}
    \end{subfigure}
    \begin{subfigure}[b]{0.19\textwidth}
    \centering
    \hspace{-1.6mm}
    \includegraphics[width=\textwidth]{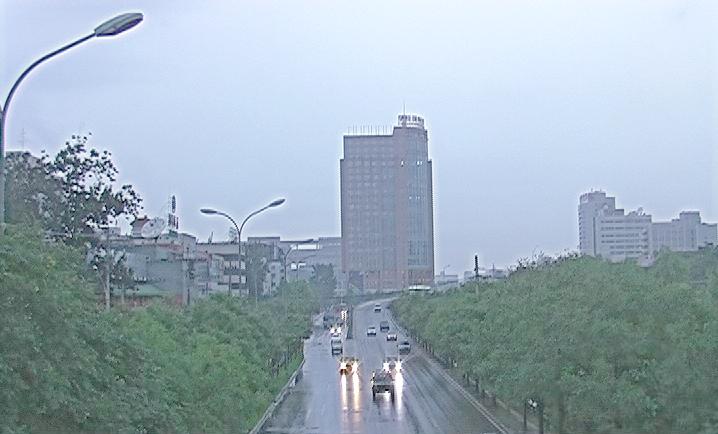}
        \caption{ZeroDCE~\cite{Zero-DCE}}
    \end{subfigure}
    \begin{subfigure}[b]{0.19\textwidth}
    \centering
        \includegraphics[width=\textwidth]{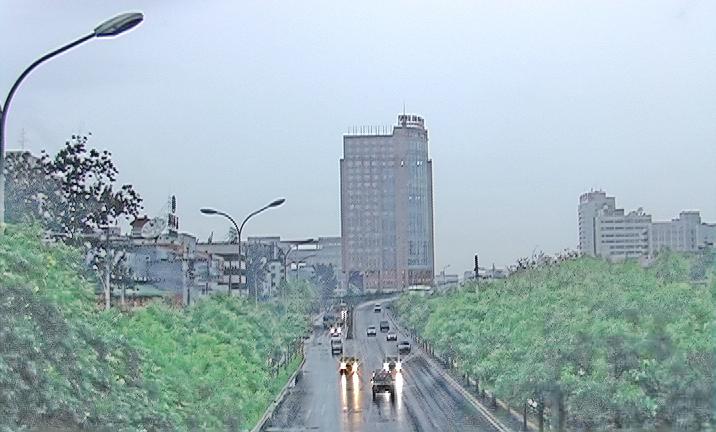}
        \caption{KinD++~\cite{kind++}}
    \end{subfigure}
    \begin{subfigure}[b]{0.19\textwidth}
        \centering
        \includegraphics[width=\textwidth]{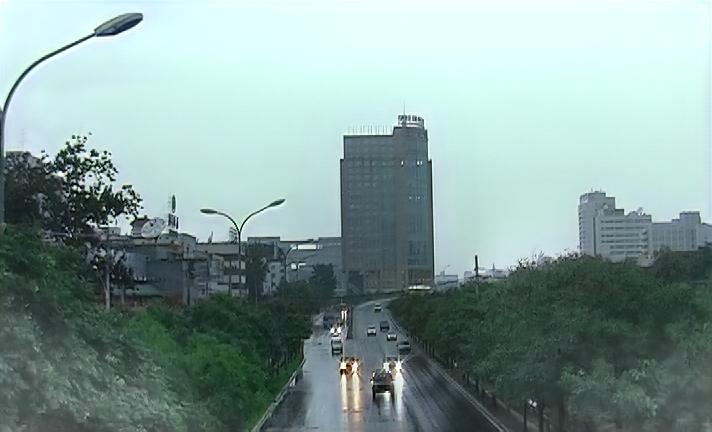}
        \caption{DRBN stage1~\cite{drbn}}
    \end{subfigure}
    \begin{subfigure}[b]{0.19\textwidth}
        \centering
        \includegraphics[width=\textwidth]{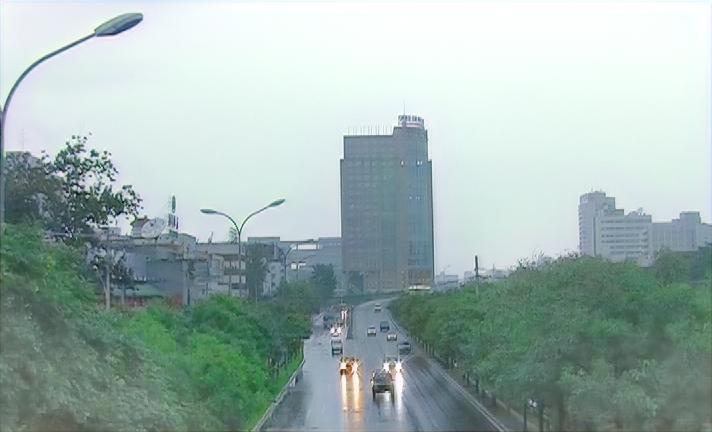}
        \caption{DRBN stage2~\cite{drbn}}
    \end{subfigure}
    \begin{subfigure}[b]{0.19\textwidth}
        \centering
        \includegraphics[width=\textwidth]{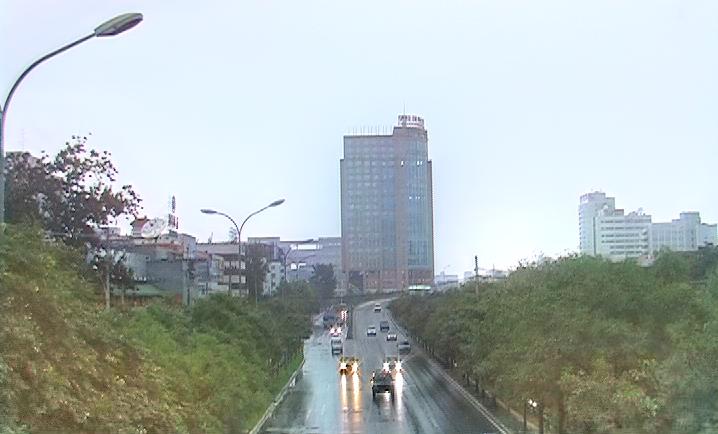}
        \caption{EnlightenGAN~\cite{jiang2021enlightengan}}
    \end{subfigure}
    \begin{subfigure}[b]{0.19\textwidth}
    \centering
        \includegraphics[width=\textwidth]{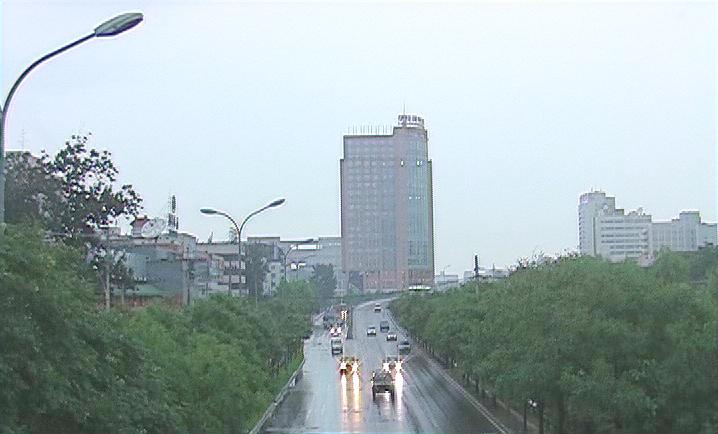}
        \caption{Inv-EnNet (Ours)}
    \end{subfigure}
    \vspace{-2mm}
    \caption{Visual comparison with SOTA methods on the NPE dataset.}
    \vspace{-2mm}
    \label{fig:sota2}
\end{figure*}

\bibliographystyle{IEEEbib}
\bibliography{Inv-EnNet}